\documentclass[lettersize,journal]{IEEEtran}
\usepackage{amsmath,amsfonts}
\usepackage{algorithmicx}
\usepackage{algorithm}
\usepackage{array}
\usepackage[caption=false,font=footnotesize,labelfont=rm,textfont=rm]{subfig}
\usepackage{textcomp}
\usepackage{caption}
\usepackage{stfloats}
\usepackage{url}
\usepackage{verbatim}
\usepackage{xcolor}
\usepackage{colortbl}
\usepackage{graphicx}
\usepackage{cite}
\usepackage{booktabs}
\usepackage{algpseudocode}
\definecolor{cvprblue}{rgb}{0.21,0.49,0.74}
\usepackage[pagebackref,breaklinks,colorlinks,citecolor=cvprblue]{hyperref}
\usepackage{multirow}

\definecolor{Gray}{gray}{0.85}

\newcommand{\method}{M$\text{P}^2$A}

\newcommand{\raya}[1]{\renewcommand{\arraystretch}{#1}}

\begin{document}

\title{Multi-Prompt Progressive Alignment for Multi-Source Unsupervised Domain Adaptation}

\author{Haoran Chen,
        Zexiao Wang, 
        Haidong Cao,
        Zuxuan Wu,
        and Yu-Gang Jiang ~\IEEEmembership{Fellow,~IEEE}% <-this % stops a space
\IEEEcompsocitemizethanks{\IEEEcompsocthanksitem H. Chen, Z. Wang, H. Cao, Z. Wu, Y-G. Jiang are with the Institute of Trustworthy Embodied AI, Fudan University. \protect  E-mail: \{chenhran21, \mbox{zexiaowang25}, hdcao24\}@m.fudan.edu.cn \{zxwu, ygj\}@fudan.edu.cn 
}}

% The paper headers
\markboth{Journal of \LaTeX\ Class Files,~Vol.~14, No.~8, August~2021}%
{Shell \MakeLowercase{\textit{et al.}}: A Sample Article Using IEEEtran.cls for IEEE Journals}

\maketitle

\begin{abstract}
Large Vision-Language Models like CLIP have become a powerful foundation for Unsupervised Domain Adaptation due to their strong zero-shot generalization. State-of-the-art methods typically leverage CLIP to generate pseudo-labels for the target domain, then fine-tune the model to learn domain-invariant features. However, these methods attempt to align source and target domains using all pseudo-labeled data simultaneously. This one-shot alignment struggles with noisy, hard-to-classify samples, leading to error propagation and suboptimal feature learning. The problem is even more amplified in the multi-source scenario, where diverse domain gaps and varying noise levels across multiple source domains further destabilize the alignment process. To address this issue, in this work, we propose a progressive alignment strategy for adapting CLIP to unlabeled downstream task. Our method begins by training the model on a high-confidence subset of target samples, allowing it to first learn a well-aligned representation from the most reliable data. As training progresses, it gradually incorporates more challenging samples, guiding the model to refine its understanding without being overwhelmed by initial label noise. This progressive approach effectively mitigates confirmation bias and promotes a more robust convergence, allowing for the learning of genuinely domain-invariant features. We name our approach \textbf{M$\text{P}^2$A} and test it on three popular UDA benchmarks, namely ImageCLEF, Office-Home, and the most challenging DomainNet. Experiments showcase that \textbf{M$\text{P}^2$A} achieves state-of-the-art performance when compared with most recent CLIP-based MS-UDA approaches, demonstrating the effectiveness of our approach.

\end{abstract}


\section{Introduction}
\IEEEPARstart{R}{ecent} years have witnessed a massive explosion in the capabilities of artificial intelligence\cite{devlin2019bert,brown2020gpt,bommasani2021opportunities}, driven by the advent of large-scale models pre-trained on vast, web-scale datasets. These models have demonstrated remarkable abilities to generalize across a wide variety of tasks, often achieving human-level performance when the test data closely resembles their training distribution. However, when deployed in real-world scenarios, these systems frequently encounter data from new environments, sensors, or styles, a phenomenon known as domain shift\cite{quinonero2008datasetshift,torralba2011unbiased,zhang2013shift}. This distributional mismatch between training and deployment data can cause a drastic and unexpected drop in accuracy\cite{de2021continualsurvey, chenarc, wang2022l2p}. While one might consider labeling deployment data for downstream adaptation, the manual annotation of data for every new domain is both prohibitively expensive and unscalable. As such, enabling models to autonomously adapt to new, unlabeled domains has become a central and fundamental challenge.

To address this challenge, Unsupervised Domain Adaptation (UDA) has emerged as a key research area, aiming to transfer knowledge from a labeled source domain to an unlabeled target domain\cite{pan2009transferlearning,csurka2017dasurvey,wang2018dasurvey}. A more challenging yet realistic extension of this setting is Multi-Source Unsupervised Domain Adaptation (MS-UDA)\cite{xu2018deep}, where labeled data from multiple, heterogeneous source domains is available for transfer. This scenario is particularly relevant as real-world data is often aggregated from a variety of sources\cite{wang2018dasurvey, singhal2023dasurvey}. However, the multi-source setting introduces significant complexities: the model must now bridge multiple, potentially conflicting domain gaps and handle varying data quality across sources. This can lead to an unstable alignment process and the risk of negative transfer, where a poorly matched source domain hinders rather than helps adaptation to the target.

\begin{figure*}
\centering
  \includegraphics[width=0.9\linewidth]{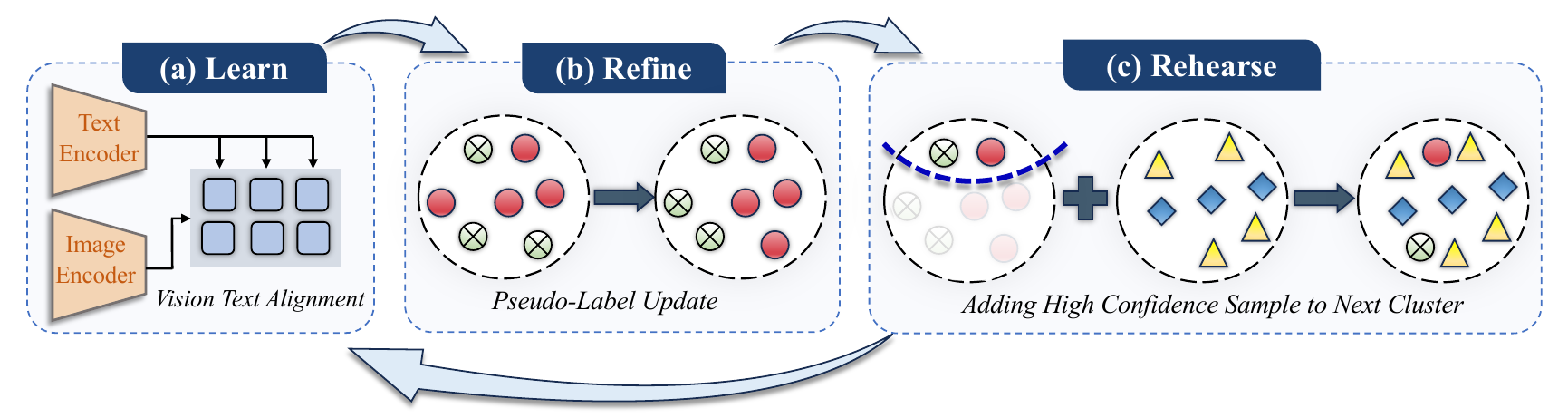}  
  \caption{Illustration of our progressive domain alignment pipeline. The adaptation process starts with the \textbf{\textit{learning}} phase, where the model is trained on the current cluster. Following this, we \textbf{\textit{refine}} the pseudo-labels of the current cluster using the updated model. As training advances, increasingly challenging classes are gradually introduced. A subset of high-confidence samples from the refined cluster is then \textbf{\textit{rehearsed}}, i.e., transferred into the next cluster’s training set. This progressive learning strategy enables stable adaptation from easy to hard samples while mitigating error accumulation and catastrophic forgetting.
}
    \label{fig:teaser}
    
\end{figure*}

Recently, researchers have leveraged large Vision-Language Models such as CLIP\cite{radford2021learning} to address the MS-UDA problem\cite{ge2023dapl, chen2023mpa, du2024damp}. With its strong zero-shot capability, CLIP can generate pseudo-labels for unlabeled target data by aligning images with text-based class descriptions. A common strategy then is to fine-tune the model on the entire pseudo-labeled target set, aiming to align the source and target feature distributions. However, this ``one-shot” alignment approach has a critical drawback: it treats all pseudo-labels with equal importance. Noisy predictions on ambiguous or hard samples are immediately incorporated into training, resulting in confirmation bias, error propagation, and suboptimal feature learning\cite{arazo2020confirmationbias}. While some studies explore iterative pseudo-label refinement through self-training\cite{mei2020st1,xie2020st2,zou2018st3}, these methods are computationally expensive, as each round processes the full training set. This challenge is further exacerbated in MS-UDA, where the compounded domain shifts and varying noise levels across multiple source domains make the initial alignment process even less reliable, demanding both greater robustness and computational efficiency.

To address the limitations of existing methods, we propose \textbf{M}ulti-Source \textbf{P}rogressive \textbf{P}rompt \textbf{A}lignment (\method), a simple yet effective framework built on a “learn, refine, and rehearse” cycle that progressively adapts CLIP to the target domain through an easy-to-hard curriculum. This cycle is designed to stabilize learning by starting from reliable supervision and gradually incorporating more ambiguous data.

Before detailing the learning cycle, we perform several key preparatory steps. First, we structure the problem space using a balanced KMeans\cite{bradley2000constrained,malinen2014balanced} algorithm to cluster classes based on their CLIP visual embeddings. To estimate the difficulty of each cluster, we compute the average cosine similarity between the image embeddings and their corresponding text-based class prototypes. Clusters with higher average similarity are deemed easier, enabling us to sort the clusters from easy to hard and establish a progressive training schedule.

Additionally, unlike prior CLIP-based UDA approaches that directly adopt zero-shot pseudo-labels from the raw CLIP model\cite{ge2023dapl,bai2024pda,du2024damp}, we argue that this overlooks valuable information from the labeled source domains, particularly in the multi-source setting. To overcome this, we first pre-train CLIP-based models individually on each source domain using supervised data. We then generate pseudo-labels on the target domain by ensembling the predictions from all source-specific models, employing strategies such as confidence averaging or majority voting to produce more reliable pseudo-labels for downstream training.

Within each curriculum stage, we begin with the learning phase, where we independently optimize a set of learnable textual prompts\cite{zhou2022coop,zhou2022cocoop} and a shared set of lightweight visual PEFT\cite{houlsby2019adapter,zhang2021tipadapter,hu2022lora} modules for each source–target domain pair. These components allow for efficient and expressive domain adaptation in both modalities. To consolidate and align the learned textual prompts, we follow prior work\cite{chen2023mpa} by passing them through a shallow auto-encoder, which projects the high-dimensional prompt vectors into a shared, denoised latent space\cite{deng2014autoencoder,zhou2022cocoop}. To further encourage consistency across source domains, we introduce an $\mathcal{L}_1$ regularization loss on the predictions from different prompt sets, guiding them toward a unified representation for the target domain.

The core innovation of our approach lies in the \textit{refine and rehearse} phase. After training on a given cluster, the model regenerates pseudo-labels for that cluster, reflecting its improved understanding. From these, we extract only the most confident samples, i.e., those exceeding a predefined probability threshold, and incorporate them into the training set for the next, more challenging cluster. This selective rehearsal mechanism serves a dual purpose: it anchors future learning with reliable examples to prevent catastrophic forgetting\cite{de2021continualsurvey}, while also constraining the training set size for improved computational efficiency. Crucially, in contrast to conventional self-training methods that indiscriminately reuse all past data\cite{xie2020st2}, our confidence-aware strategy reduces error propagation and mitigates overfitting to noisy labels. By progressively expanding the training data with only well-aligned samples, \method~achieves more stable convergence and learns domain-invariant representations in a robust and scalable manner. The whole pipeline of our training cycle is illustrated in Figure~\ref{fig:teaser}.

We validate the effectiveness of our proposed framework through extensive experiments on three widely-used MS-UDA benchmarks: ImageCLEF~\cite{caputo2014imageclef}, Office-Home~\cite{venkateswara2017officehome}, and DomainNet~\cite{peng2019m3sda}. These datasets encompass a broad range of domains, visual styles, and adaptation challenges, providing a comprehensive testbed for evaluating multi-source domain adaptation methods. Across all benchmarks, \method~achieves average accuracies of 94.3\%, 91.8\%, and 64.1\%, respectively, consistently surpassing recent CLIP-based UDA approaches by a notable margin. These results demonstrate the robustness, scalability, and generalization capability of our progressive alignment framework, particularly under the challenging multi-source scenarios.

A preliminary version of this work, MPA, appeared in~\cite{chen2023mpa}, where we were among the first to explore the use of CLIP for domain adaptation. Since then, the research landscape has evolved rapidly, with numerous follow-up studies advancing CLIP-based adaptation\cite{du2024damp, singha2023adclip}. To remain competitive and push the frontier further, this journal version introduces several substantial improvements over the original conference paper. 

First, while the earlier version primarily emphasized textual prompt learning, the current work fully exploits both the textual and visual modalities of CLIP by incorporating PEFT modules. This enables the model to learn more expressive and adaptable visual representations. Second, we introduce a novel source pretraining strategy that leverages the supervision from labeled source domains to generate more robust and accurate pseudo-labels for the target domain. Third, we propose a curriculum-based alignment framework that progressively adapts the model from easy to hard samples, effectively mitigating confirmation bias and improving training stability. 

In addition, we significantly expand the experimental scope: we re-implement and fairly compare our method against a comprehensive set of strong CLIP-based baselines using modern ViT-based backbones, aligning with current trends in large-scale vision models. Finally, we provide in-depth empirical analysis to validate the robustness and compatibility of our proposed progressive alignment strategy across diverse domain adaptation benchmarks.

\section{Related Works}
\subsection{Multi-Source Unsupervised Domain Adaptation}

First introduced by Yang et al.\cite{yang2007cross}, MS-UDA extends the standard UDA setting by leveraging multiple labeled source domains to improve generalization to an unlabeled target domain. Classic approaches often focus on aligning distributions across domains. Adversarial methods like MDAN\cite{zhao2018adversarial} and DCTN \cite{xu2018deep} train domain discriminators to encourage feature alignment, while discrepancy-based methods such as MFSAN\cite{zhu2019aligning} and $\text{M}^3$SDA \cite{peng2019m3sda} align each source-target pair by minimizing statistical distances like maximum mean discrepancy or moment distance.

Due to the emerging of large-scale pretrained networks, self-training has become a dominant strategy in UDA, where pseudo-labels generated on target data are used to iteratively refine the model. For example, CST\cite{liu2021cst} employs a bi-model framework where two networks alternately generate pseudo-labels for each other, reducing overfitting to self-predictions. Extending this to the multi-source case, CSR\cite{zhou2024csr} introduces an instance-level ensemble of source-specific models and a domain-ensemble network, refined in a mutual loop via cross-model supervision. These self-training strategies demonstrate the effectiveness of leveraging target structure and inter-model consistency to enhance multi-source adaptation.

\subsection{CLIP based MS-UDA}

Recent advances in vision-language models\cite{jia2021scaling,li2022blip,alayrac2022flamingo}, especially CLIP\cite{radford2021learning}, have sparked new research in domain adaptation by leveraging rich multimodal semantics. A pioneering effort in this direction is DAPL\cite{ge2023dapl}, which introduces prompt learning into UDA by disentangling domain and category representations. It optimizes domain-specific prompts to maintain semantic structures while avoiding the limitations of hard domain alignment. Building on this idea, subsequent works have proposed various strategies to enhance prompt learning. AD-CLIP\cite{singha2023adclip} and DAMP\cite{du2024damp} extend this direction by designing domain-invariant prompts and integrating visual context into the prompting process. PDA\cite{bai2024pda} further incorporates domain knowledge into a two-branch prompt tuning framework, aligning class and domain representations via feature banks and contrastive losses. However Most existing CLIP-based UDA methods, are limited to the single-source setting. Addressing this gap, MPA\cite{chen2023mpa} is the first to tackle multi-source UDA, introducing a lightweight and modular multi-prompt alignment strategy that learns separate prompts per source-target pair and aligns them via autoencoding and consensus optimization. Inspired by these insights, PGA\cite{phan2024pga} formulates UDA as a multi-objective optimization problem and aligns prompt gradients across domains to encourage consistent learning, achieving robust adaptation under both single and multi-source settings. Collectively, these methods demonstrate the promise of prompt-based CLIP adaptation in both efficiency and accuracy for domain adaptation.

\subsection{Curriculum Learning}

The concept of Curriculum Learning, formally introduced by~\cite{bengio2009curriculum}, advocates for training models on a structured sequence of examples from easy to hard. This paradigm, inspired by human learning, aims to guide optimization towards more robust solutions and improve generalization by first building a foundation on simpler concepts\cite{Zhang_2017_ICCV,zhou2018minimax,Chen_2019_CVPR}.

Despite its potential, curriculum learning remains underutilized in UDA. One notable effort is C-SFDA\cite{karim2023cfsda}, which applies a dynamic, self-paced curriculum based on prediction confidence and uncertainty. To determine difficulty, C-SFDA measures variance across multiple test-time augmentations and adaptively thresholds sample reliability. While effective, this process requires a substantial amount of hand-tuned scheduling and heuristic design including augmentation policies, uncertainty statistics, and carefully calibrated thresholds, making the method complex and sensitive to hyperparameter settings.

In contrast, our approach introduces a static, data-driven curriculum that is both simpler and more robust. We predefine the learning schedule by clustering classes using CLIP visual embeddings and ranking them by image-text alignment difficulty, independent of model predictions. Our "learn, refine, and rehearse" framework progressively introduces harder classes while anchoring training with high-confidence examples from earlier stages. This avoids the need for online confidence computation and extensive scheduling logic, resulting in a more interpretable and computationally efficient curriculum learning strategy.

\begin{figure*}
    \centering
   \includegraphics[width=\linewidth]{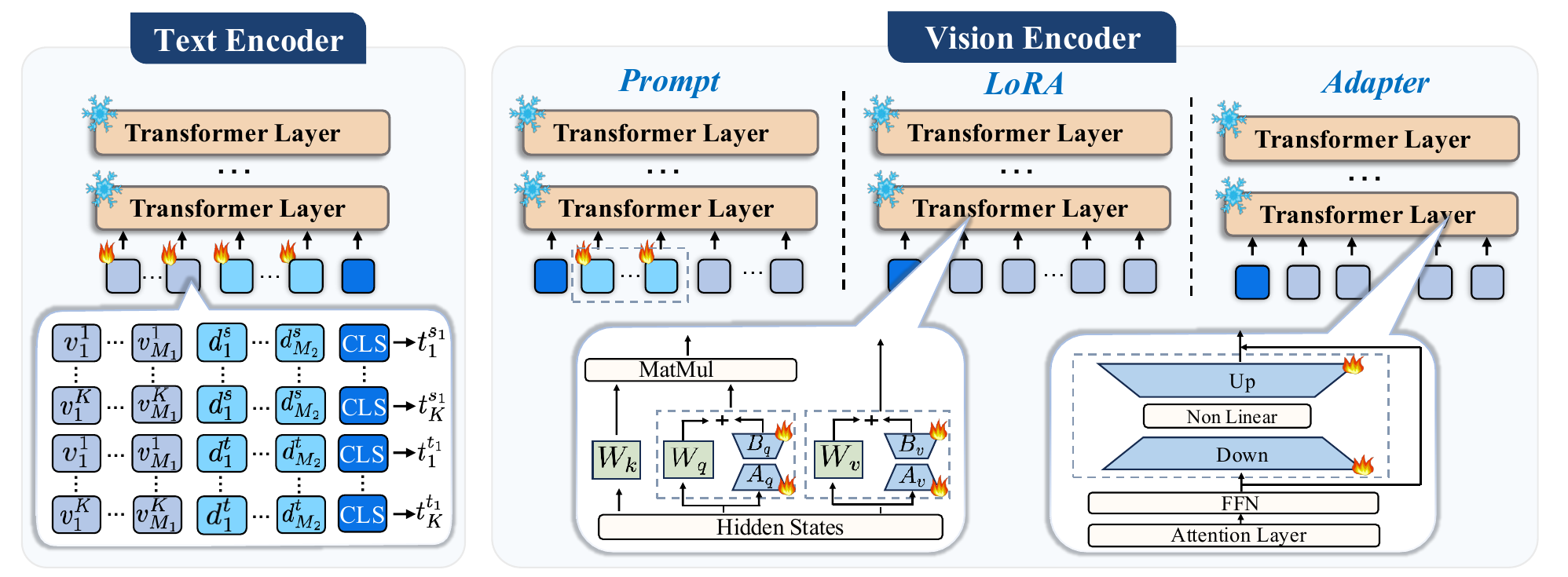}
   \caption{
Illustration of our architectural design. The \textbf{left} panel shows the text encoder, where textual prompts consist of class-specific context tokens $v_i^k$ and domain-specific tokens $d_j^d$. Only these prompt tokens are updated during training, while the rest of the text encoder remains frozen. The \textbf{right} panel presents three visual PEFT modules used to adapt the CLIP visual encoder to the target domain: (1) \textbf{Visual Prompt}, where learnable tokens are inserted at the input of each transformer\cite{vaswani2017transformer,dosovitskiy2020vit}  layer; (2) \textbf{LoRA}, where trainable low-rank matrices $A$ and $B$ are merged with the query and value projections during self-attention; and (3) \textbf{Adapter}, where lightweight adapter blocks with down-projection, non-linearity, and up-projection are inserted after the feed-forward network.
}
\label{fig:architecture}
\end{figure*}

\section{Method}
To achieve more robust alignment on the target domain, we propose a simple yet effective ``learn, refine, and rehearse” framework built upon the pre-trained CLIP model, leveraging its strong zero-shot capability. The core idea is to partition the learning data into subsets and train the model in a sequential, curriculum-based manner that progresses from easy to hard examples. In this way, the adaptation is gradually enhanced and the impact of noisy supervision is mitigated. In the following sections, we begin by introducing the design of textual prompts and visual parameter-efficient tuning modules in Section~\ref{sec:architectural design}. The procedure for generating initial pseudo-labels is then detailed in Section~\ref{sec:initial pseudo label}. Next, we describe our data partitioning strategy based on CLIP embeddings in Section~\ref{sec:curriculum grouping}. The core learning phase is presented in Section~\ref{sec:cluster learning phase}, followed by the refine-and-rehearse phase in Section~\ref{sec:refine and rehearse phase}.

\subsection{Architectural design}
\label{sec:architectural design}

Let $N$ denote the total number of domains, where the first $N{-}1$ domains are source domains and the $N$-th domain is the target domain. Let $K$ denote the number of classes shared across all domains. In the multi-source UDA setting, our objective is to learn a domain-invariant latent space such that both the inter-source domain gaps and the discrepancies between each source and the target domain are minimized. To this end, we design textual prompts that incorporate both domain-invariant and domain-specific features, and introduce visual Parameter-Efficient Fine-Tuning modules to enhance visual adaptation to the target domain.

\subsubsection{\textbf{Textual Prompt Design}}
For the textual prompts, we construct two types of learnable tokens: class-specific context tokens $v_i^k$, where $i \in \{1, 2, \ldots, M_1\}$ and $k \in \{1, 2, \ldots, K\}$, and domain-specific tokens shared across all classes, denoted as $d_j^d$, where $j \in \{1, 2, \ldots, M_2\}$ and $d \in \{s, t\}$. Here, $M_1$ and $M_2$ represent the number of tokens in each category, and $s$ and $t$ indicates source and target domains respectively. Combining these tokens, we construct prompts for $2K$ categories (source and target for each class) used in contrastive training. Specifically, for each source-target pair, we define the prompt embedding matrix as:
\[
P_i = [t_1^{s_i}, \ldots, t_K^{s_i},\ t_1^{t_i}, \ldots, t_K^{t_i}]^\top, \quad i \in \{1, 2, \ldots, N{-}1\}.
\]
An illustration of this is shown in the left of Figure~\ref{fig:architecture}.

\subsubsection{\textbf{Visual PEFT Modules}}
The original MPA framework utilizes the raw visual embeddings directly from the frozen CLIP vision encoder. In contrast, for \method~we take a step further and adapt the visual encoder to the target domain in a parameter-efficient manner. Specifically, we investigate three lightweight fine-tuning strategies: prompt tuning\cite{jia2022vpt}, LoRA\cite{hu2022lora}, and adapter\cite{houlsby2019adapter} modules. These modules are inserted into each Transformer block, and only their parameters are updated, while the rest of the vision encoder remains frozen.

Prompt tuning prepends a set of learnable prompt tokens to the patch embeddings at each Transformer layer. Given an input image represented as patch embeddings $X \in \mathbb{R}^{n \times d}$, where $n$ is the number of patches and $d$ is the embedding dimension, we introduce learnable tokens $E \in \mathbb{R}^{M_3 \times d}$ and modify the input as:
\[
\tilde{X} = [E; X] \in \mathbb{R}^{(M_3+n) \times d}
\]

LoRA modifies the weight matrices in the attention layers by injecting low-rank updates. For a given weight matrix $W \in \mathbb{R}^{d \times d}$, two trainable matrices $A \in \mathbb{R}^{r_1 \times d}$ and $B \in \mathbb{R}^{d \times r_1}$ with $r_1 \ll d$ are introduced, and the updated matrix becomes:
\[
W' = W + \Delta W, \quad \Delta W = BA
\]
In our design, LoRA is applied to the query and value projection matrices in the self-attention mechanism:
\[
Q = (W_q + B_q A_q) X, \quad V = (W_v + B_v A_v) X
\]

Adapter modules are lightweight residual layers inserted within each Transformer block. A typical adapter is composed of a down-projection, a nonlinearity, and an up-projection:
\[
\text{Adapter}(h) = h + W_{\text{down}}(\text{ReLU}(W_{\text{up}} h))
\]
where $W_{\text{up}} \in \mathbb{R}^{d \times r_2}$ and $W_{\text{down}} \in \mathbb{R}^{r_2 \times d}$ with $r_2 \ll d$. The adapter output is added back residually and layer normalized:
\[
h' = \text{LayerNorm}(h + \text{Adapter}(h))
\]

These visual PEFT modules are designed to guide the image feature extraction process at different hierarchical levels, enabling the model to focus more on task-relevant semantic regions and thereby generate more discriminative visual features. It is worth noting that, for simplicity, we employ a single shared set of visual PEFT modules across all source–target domain pairs.

\subsection{Initial Pseudo-Label Generation}
\label{sec:initial pseudo label}

Most existing CLIP-based domain adaptation methods generate pseudo-labels by directly applying CLIP’s zero-shot predictions on the target domain. However, this approach fails to leverage the rich semantic knowledge embedded in the labeled source domains, which can provide valuable guidance for more accurate pseudo-label generation.

To address this limitation, we propose a supervised ensemble-based strategy that exploits the source domain annotations. Specifically, for each source domain $\mathcal{D}_s^i$ where $i \in \{1, 2, \ldots, N{-}1\}$, we fine-tune the CLIP model using the textual prompts and visual PEFT modules described in Section~\ref{sec:architectural design}, leveraging only the labeled data from that source. This results in $N{-}1$ independently trained models. Each model is then used to generate class predictions on the unlabeled target domain samples, producing a set of logits:
\[
\{\mathbf{z}_i(x_t)\}_{i=1}^{N-1}, \quad \text{for each } x_t \in \mathcal{D}_t.
\]

To aggregate these predictions into a final pseudo-label $\hat{y}_t$ for each target sample $x_t$, we consider two voting strategies:

\begin{itemize}
    \item \textbf{Average Confidence Voting}: Average the softmax outputs across all models and select the most confident class:
    \[
    \hat{y}_t = \arg\max_k\ \frac{1}{N-1} \sum_{i=1}^{N-1} \text{softmax}(\mathbf{z}_i(x_t))_k.
    \]

    \item \textbf{Majority Voting}: Take the class that receives the most votes from the $N{-}1$ models:
    \[
    \hat{y}_t = \text{mode} \left( \{\arg\max_k\ \mathbf{z}_i(x_t)_k\}_{i=1}^{N-1} \right).
    \]
\end{itemize}

The choice of voting strategy is selected empirically. Importantly, since the source-trained models are prone to overfitting their respective source domains, they are not used in the subsequent adaptation process. That is, this training step is performed solely to provide a more informed and stable initialization of pseudo-labels for the target domain.

\subsection{Curriculum-Oriented Data Grouping}
\label{sec:curriculum grouping}

Traditional approaches in unsupervised domain adaptation often align source and target domains using the entire dataset at once. However, this practice can be suboptimal due to the presence of noisy or hard-to-classify target samples, particularly when pseudo-labels are involved. Noisy labels may introduce confirmation bias and hinder model convergence. To address this, we propose a curriculum-style strategy that partitions the data into subsets and performs alignment progressively, starting from easier samples and moving toward more difficult ones.

We begin by generating pseudo-labels $\{\hat{y}_t\}_{t=1}^{|\mathcal{D}_t|}$ for the unlabeled target samples $\mathcal{D}_t = \{x_t^{(1)}, x_t^{(2)}, \ldots, x_t^{(n)}\}$ using the method described in Section~\ref{sec:initial pseudo label}. For each class $k \in \{1, 2, \ldots, K\}$, we compute its class-wise feature centroid $\mathbf{c}_k$ as the average of CLIP visual embeddings $\phi(x)$ for target samples assigned to class $k$, where $\phi(\cdot)$ denotes the CLIP vision encoder:
\[
\mathbf{c}_k = \frac{1}{|\mathcal{X}_k|} \sum_{x \in \mathcal{X}_k} \phi(x), \quad \mathcal{X}_k = \{x \in \mathcal{D}_t \mid \hat{y}_t = k\}.
\]

Next, we apply balanced K-Means clustering to the set of class centroids $\{\mathbf{c}_1, \mathbf{c}_2, \ldots, \mathbf{c}_K\}$ to partition them into $T$ clusters $\{\mathcal{C}_1, \ldots, \mathcal{C}_T\}$. Unlike standard K-Means, the balanced variant enforces that each cluster contains approximately the same number of class centroids. This is achieved by initializing cluster centers using standard K-Means and then reassigning centroids while enforcing the size constraint. If a centroid’s nearest cluster has already reached the target capacity, it is assigned to the next nearest available cluster with space. The cluster centers are updated after each full assignment round, and the process is repeated until convergence. The process is presented in Algorithm~\ref{alg:balanced_kmeans}.

Once the class centroids are grouped, we assess the difficulty of each cluster. For each class $k$ in a cluster $\mathcal{C}_t$, we compute the cosine similarity between its visual embedding $\phi(x)$ and its corresponding textual embedding $\psi(t_k)$. Here $\psi(\cdot)$ denotes the CLIP text encoder. The average similarity score across all classes in a cluster serves as a proxy for how confidently CLIP aligns images and text for those classes:
\[
s_t = \frac{1}{|\mathcal{C}_t|} \sum_{k \in \mathcal{C}_t} \frac{1}{|\mathcal{X}_k|} \sum_{x \in \mathcal{X}_k} \cos(\phi(x), \psi(t_k)).
\]

We then sort the clusters $\{\mathcal{C}_1, \ldots, \mathcal{C}_T\}$ in descending order of $s_t$ to form a progressive curriculum from easy to hard, as clusters with higher average similarity scores indicates that CLIP is already more confident about their alignment. 

\begin{algorithm}
\caption{Balanced K-Means Clustering}
\label{alg:balanced_kmeans}
\begin{algorithmic}[1]
\Require Feature set $\mathcal{C} = \{\mathbf{c}_1, \ldots, \mathbf{c}_K\}$, cluster number $T$
\Ensure Cluster assignments $\{\mathcal{C}_1, \ldots, \mathcal{C}_T\}$

\State Initialize centroids $\{\mu_1, \ldots, \mu_T\}$ using K-Means
\State Set cluster size budget $B \gets \lceil K/T \rceil$
\Repeat
\State Initialize empty clusters: $\mathcal{C}_1 \gets \emptyset, \ldots, \mathcal{C}_T \gets \emptyset$
    \ForAll{$\mathbf{c}_k \in \mathcal{C}$}
        \State Compute distances $d_j = \|\mathbf{c}_k - \mu_j\|_2$
        \State Sort clusters by increasing $d_j$
        \ForAll{cluster $j$ in sorted order}
            \If{$|\mathcal{C}_j| < B$}
                \State Assign $\mathbf{c}_k$ to $\mathcal{C}_j$
                \State \textbf{break}
            \EndIf
        \EndFor
    \EndFor
    \For{$j = 1$ to $T$}
        \State Update $\mu_j \gets \frac{1}{|\mathcal{C}_j|} \sum_{\mathbf{c} \in \mathcal{C}_j} \mathbf{c}$
    \EndFor
\Until{convergence}
\end{algorithmic}
\end{algorithm}

\begin{figure*}
    \centering
   \includegraphics[width=0.9\linewidth]{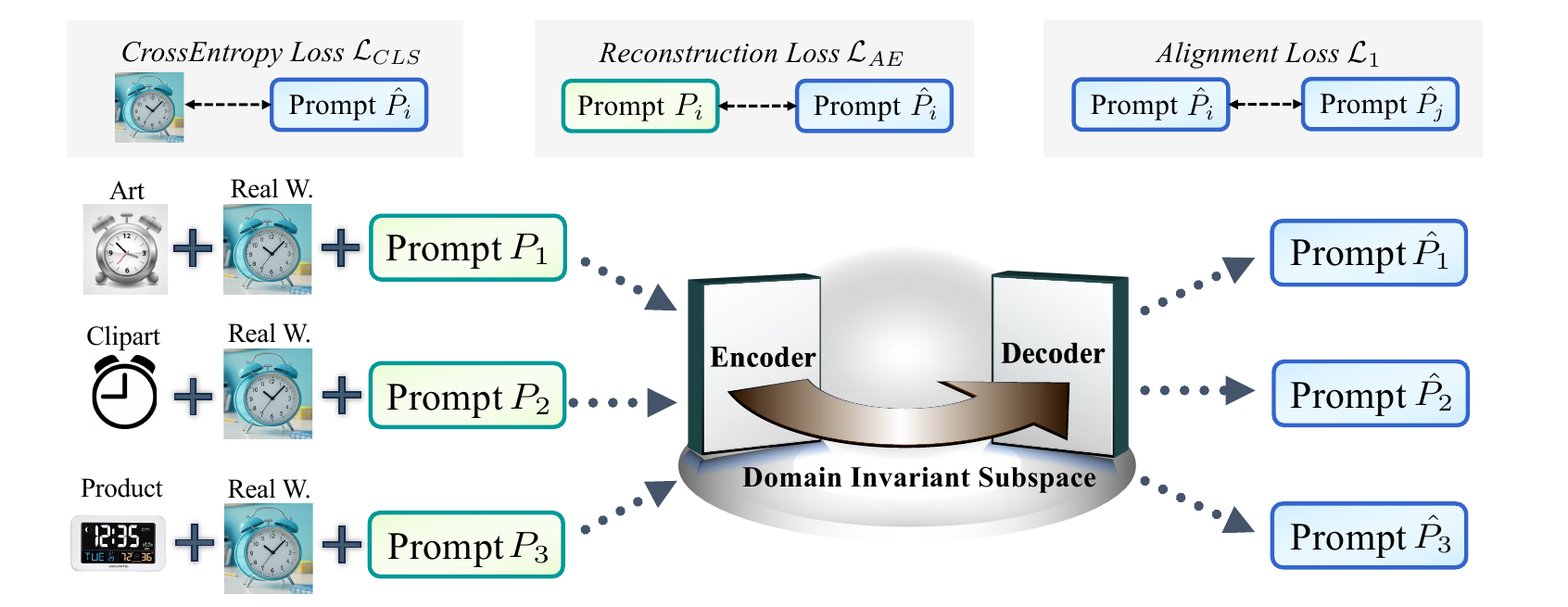}
\caption{Example of the Multi-Prompt Alignment phase on the Office-Home dataset. Here, $P_1$, $P_2$, and $P_3$ denote prompts obtained from the Initial Feature Adaptation phase for the Art–Real World, Clipart–Real World, and Product–Real World pairs, respectively. All prompts are projected into a domain-invariant subspace for alignment using an auto-encoder architecture. During this phase, we apply a cross-entropy loss on pseudo-labeled target samples, a reconstruction loss to preserve information between the original and reconstructed prompts, and an alignment loss across all generated prompts.
}
\label{fig:method}

\end{figure*}

\subsection{Cluster Learning Phase}
\label{sec:cluster learning phase}
With all components in place, we now describe how each cluster is progressively trained. Following the MPA framework, this process is divided into two steps: an initial feature adaptation step and a combined prompt alignment step.

\subsubsection{\textbf{Initial Feature Adaptation}} 
To enable better adaptation to the target domain, we train individual sets of textual prompts for each source–target domain pair. In addition, we jointly optimize a shared visual PEFT module across all domains. To reduce the impact of noisy pseudo-labels, we apply a confidence threshold $\tau$ and include only target samples with predicted class probability greater than $\tau$ during training. Let $P_i$ denote the textual prompts for the $i$-th source–target pair, where $i \in \{1, 2, \ldots, N-1\}$, and let $\mathcal{V}$ represent the shared visual PEFT module. Given a labeled image $x_s$ from a source domain $\mathcal{D}_s$ with label $y_s$, and an unlabeled image $x_t$ from the target domain $\mathcal{D}_t$ with pseudo-label $\hat{y}_t$, the optimization objective is defined as:

\begin{equation}
\resizebox{\columnwidth}{!}{
$\displaystyle
\min_{P_i, \mathcal{V}} 
  -\frac{1}{n_s} \sum_{x^s \sim \mathcal{D}_s} \log P(y = y_s \mid x^s; P_i, \mathcal{V})
  -\frac{1}{n_t} \sum_{x^t \sim \mathcal{D}_t} \log P(y = \hat{y}_t \mid x^t; P_i, \mathcal{V})
$}
\label{eq:prompt-loss}
\end{equation}

Let $\mathcal{T}$ denote a temperature scaling value, the class probability for image $x^d$ (where $d \in \{s, t\}$) in Eqn \ref{eq:prompt-loss} is computed using softmax-normalized cosine similarity between the visual and textual embeddings:

\begin{equation}
P(y = k \mid x^d) = 
\frac{\exp(\langle \psi(t^d_k), \phi(x^d) \rangle / \mathcal{T})}{
\sum_{d' \in \{s, t\}} \sum_{i=1}^{K} \exp(\langle \psi(t^{d'}_i), \phi(x^d) \rangle / \mathcal{T})}
\label{eq:prob}
\end{equation}

\subsubsection{\textbf{Multi-Prompt Alignment}} 
The above training yields a distinct textual prompt $P_i$ for each source–target pair. To further consolidate the learned knowledge and eliminate redundancy, we perform multi-prompt alignment. Although directly using the learned prompts can yield decent results, the high-dimensional nature of prompt vectors may encode redundant or domain-specific noise\cite{wang2014generalized}. Inspired by the manifold hypothesis\cite{fefferman2016testing}, we employ a denoising autoencoder to extract compact, domain-invariant representations of the prompts.

Specifically, let $\textbf{Proj}(\cdot)$ be a projection function implemented as a one-layer feed-forward network, and let $\textbf{Proj}_b(\cdot)$ be a two-layer non-linear decoder. Each learned prompt $P_i$ is projected into a latent space:
\begin{equation}
\textbf{Proj}(P_i) = \tilde{P}_i = W_1 P_i + b_1
\label{eq:proj}
\end{equation}
and then reconstructed back to $\hat{P}_i$ via:
\begin{equation}
\textbf{Proj}_b(\tilde{P}_i) = W_3 (\tanh(W_2 \tilde{P}_i + b_2)) + b_3
\label{eq:projb}
\end{equation}

The reconstruction objective is:
\begin{equation}
\mathcal{L}_{AE} = \frac{1}{N - 1} \sum_{i=1}^{N-1} \left\| \hat{P}_i - P_i \right\|_2^2
\label{eq:ae}
\end{equation}

While this has proven to be an effective alignment strategy \cite{ilse2020diva}, differences in data volume and domain gap across sources may still result in inconsistent prompt behavior. To promote deeper alignment among prompts, we introduce an $\mathcal{L}_1$ consistency loss based on predictions over the same target sample $x_t$:

\begin{equation}
\resizebox{\columnwidth}{!}{
$\displaystyle
\mathcal{L}_1 = \frac{2}{(N - 1)(N - 2)} \sum_{j=1}^{N - 2} \sum_{i = j+1}^{N - 1} 
\left| P(y = k^t \mid x_t; P_i) - P(y = k^t \mid x_t; P_j) \right|$
}
\label{eq:l1}
\end{equation}

% \subsubsection*{Overall Objective}
Putting these together, the total training loss in this step is:
\begin{equation}
\mathcal{L} = \mathcal{L}_{CLS} + \mathcal{L}_{AE} + \alpha \mathcal{L}_1
\label{eq:loss}
\end{equation}

\noindent where $\mathcal{L}_{CLS}$ is a regular cross-entropy loss , and $\alpha$ is a hyperparameter controlling the weight of the alignment term $\mathcal{L}_1$. During this alignment stage, the visual PEFT module $\mathcal{V}$ is kept fixed; only the textual prompts and autoencoder parameters are updated. An overview of the multi-prompt learning and alignment procedure is illustrated in Fig \ref{fig:method}.

\subsection{Refine and Rehearse Phase}
\label{sec:refine and rehearse phase}

After completing the training on each curriculum stage (i.e., cluster), we refine the pseudo-labels for that stage using the updated model. Specifically, we re-evaluate all target samples in the current cluster and regenerate their pseudo-labels based on the latest model predictions. To ensure reliability and improve computational efficiency, we apply a confidence threshold $\beta$ and retain only the most confident samples—those with predicted class probabilities exceeding $\beta$.

These high-confidence samples serve as anchor points and are carried forward to the training set of the next cluster. This mechanism enables a ``refine and rehearse" strategy: refined pseudo-labels from previously learned easier clusters are used to stabilize training on subsequent harder clusters. This gradual accumulation of trusted samples helps mitigate noise, reinforce previously learned knowledge, prevent catastrophic forgetting, and promote smoother convergence. The refinement and rehearsal process continues iteratively across all $T$ clusters until the entire target dataset has been covered.

Finally, after training on all clusters is complete, we obtain the final model. During inference, we aggregate the predictions from all domain-specific prompt sets and use the averaged output as the final prediction for each test sample.

\begin{table*}[!t]
% \vskip 0.1in
\caption{Accuracy (\%) on ImageCLEF and Office-Home. We re-implement all compared approaches using the CLIP ViT-B/16 backbone for a fair comparison.}
\begin{center}
\setlength{\tabcolsep}{3pt}
   \resizebox{\linewidth}{!}{
   \raya{1.0}
\begin{tabular*}{0.8\textwidth}{@{\extracolsep{\fill}\quad}*{11}c}
\toprule
 & & \multicolumn{4}{c}{\textbf{ImageCLEF}} & \multicolumn{5}{c}{\textbf{Office-Home}}\\ 
\addlinespace[0.5ex]
 \textbf{Method}&\textbf{Venue}
 &\textbf{$\rightarrow$ C} &\textbf{$\rightarrow$ I}  &\textbf{$\rightarrow$ P} & \textbf{Avg} & \textbf{$\rightarrow$ Ar} &\textbf{$\rightarrow$ Cl}  &\textbf{$\rightarrow$ Pr} &\textbf{$\rightarrow$ Rw} & \textbf{Avg}\\
%  \cmidrule(lr){2-6} \cmidrule(lr){7-10}
 \cmidrule{1-1} \cmidrule{2-2} \cmidrule{3-6} \cmidrule{7-11}
 \multicolumn{1}{l}{\textbf{Zero-Shot}} \\
\multicolumn{1}{l}{CLIP\cite{radford2021learning}} & ICML'21 & 97.0  & 96.8  & 82.7 & 92.2 &83.4  &65.3 &89.0   &89.4  &81.8\\
 \cmidrule{1-1} \cmidrule{2-2} \cmidrule{3-6} \cmidrule{7-11}

 \multicolumn{1}{l}{\textbf{Non-CLIP Based}} \\
\multicolumn{1}{l}{ICON\cite{yue2023icon}}  & NeurIPS'23  &97.8    &97.4     &83.2   &92.8 & 88.2     &81.1     &93.5     &93.9   &89.2\\
\multicolumn{1}{l}{CSR\cite{zhou2024csr}}  & AAAI'24 & 98.3 & 95.5 & 82.0 & 91.9  &86.2 &79.2 &92.4 &91.3 & 87.3\\
 \cmidrule{1-1} \cmidrule{2-2} \cmidrule{3-6} \cmidrule{7-11}
 \multicolumn{1}{l}{\textbf{CLIP Based}} \\

\multicolumn{1}{l}{AD-CLIP\cite{singha2023adclip}} &ICCV'23   & 97.8 & 96.7 & 82.3 & 92.3 & 84.2 & 72.6 & 92.9 & 91.2 & 85.2 \\
\multicolumn{1}{l}{MPA\cite{chen2023mpa}} & NeurIPS'23  & 97.8    & 97.4   & 81.2  &  92.1 & 85.1     &76.6     &91.3     &91.2   &86.1 \\
\multicolumn{1}{l}{DAPL\cite{ge2023dapl}} & TNNLS'23  & 97.8 & 97.2 & 83.5 & 92.8 & 84.3 & 72.4 & 92.6 & 91.4 & 85.2\\
\multicolumn{1}{l}{PDA\cite{bai2024pda}} &AAAI'24  & 97.7 & 97.3 & 81.3 & 92.1    & 87.3 & 73.9 & 92.2 & 92.4 & 86.5\\
\multicolumn{1}{l}{PGA\cite{phan2024pga}}  & NeurIPS'24 & 97.4 & 96.5 & 82.5 & 92.1& 83.6 & 76.9 & 92.9 & 91.2 & 86.2\\
\multicolumn{1}{l}{DAMP\cite{du2024damp}}  & CVPR'24 & 98.5 & 97.7 & 81.8 & 92.7 & 87.3 & 77.3 & 94.1 & 92.7 & 87.8\\
 \cmidrule{1-1} \cmidrule{2-2} \cmidrule{3-6} \cmidrule{7-11}

\multicolumn{1}{l}{\textbf{\method}-Adapter}  & -  & 97.7 & 92.3 & 78.2 & 89.4 & 90.3 & 83.2 & 92.9 & 89.1 & 88.9\\
\multicolumn{1}{l}{\textbf{\method}-Prompt}  & -  &97.8 & 97.1 & 83.7 & 92.9 & 90.2 & 80.6 & \textbf{94.8} & 92.5 & 89.5 \\
\multicolumn{1}{l}{\textbf{\method}-LoRA}  & -  & \textbf{98.8} & \textbf{99.0} & \textbf{85.0} & \textbf{94.3} & \textbf{93.6} & \textbf{83.2} & \textbf{94.8} & \textbf{95.6} & \textbf{91.8} \\
\bottomrule
\end{tabular*}}

\label{tab:imageclef Office-Home result}
\end{center}
\vspace{2ex}
\label{tab:oh_clef_performance}

\end{table*}

\section{Experiments}

\subsection{Experimental Setup}
To comprehensively assess the effectiveness of the proposed \method~method, we perform experiments on three widely-used benchmark datasets in the field of MS-UDA: ImageCLEF, Office-Home, and DomainNet.
ImageCLEF is a relatively small dataset comprising 1,800 images from 12 object categories across three distinct domains: ImageNet ILSVRC 2012 (I), Pascal VOC 2012 (P), and Caltech-256 (C).
Office-Home represents a medium-scale dataset containing approximately 15,500 images from 65 categories, drawn from four diverse domains: Art, Clipart, Product, and Real-World.
In contrast, DomainNet is currently the largest and most challenging UDA benchmark, encompassing around 600,000 images spanning 345 categories over six heterogeneous domains: Clipart, Infograph, Painting, Quickdraw, Real, and Sketch.

We use top-1 accuracy as our evaluation metric and compare our method against several representative baselines, including:  (1) zero-shot CLIP inference.
(2) CLIP-based UDA approaches, including DAMP\cite{du2024damp}, DAPL\cite{ge2023dapl}, MPA\cite{chen2023mpa}, PDA\cite{bai2024pda}, PGA\cite{phan2024pga}, and AD-CLIP\cite{singha2023adclip}. Among these, PDA, DAMP, and DAPL are originally designed for single-source UDA. To adapt them to the multi-source setting, we follow a common practice of merging all source domains into a unified source domain before applying these methods. The remaining approaches are specifically developed for multi-source UDA, but they rely on ResNet-based CLIP image encoders. To ensure a fair comparison, we re-implement all methods using the pre-trained ViT-B/16 image encoder from CLIP.
(3) Non-CLIP-based UDA methods, including ICON\cite{yue2023icon} and CSR\cite{zhou2024csr}. For consistency, we also replace their original image encoders with the CLIP ViT-B/16 backbone.

\begin{table*}[!t]
\caption{Accuracy (\%) on DomainNet. We re-implement all compared approaches using the CLIP ViT-B/16 backbone for a fair comparison.}
\vspace{1ex}
\begin{center}
\setlength{\tabcolsep}{3pt}
   \resizebox{0.9\linewidth}{!}{
    \raya{1.0}
   %  % let TeX compute the intercolumn space
\begin{tabular*}{0.7\textwidth}{@{\extracolsep{\fill}\quad}*{9}c}
\toprule
 & & \multicolumn{7}{c}{\textbf{DomainNet}} \\ 
\addlinespace[0.5ex]
 \textbf{Method} & \textbf{Venue} &\textbf{$\rightarrow$ Clp} &\textbf{$\rightarrow$ Inf}  &\textbf{$\rightarrow$ Pnt} &\textbf{$\rightarrow$ Qdr} & \textbf{$\rightarrow$ Rel}  & \textbf{$\rightarrow$ Skt}  &\textbf{Avg}\\
  \cmidrule{1-1} \cmidrule{2-2} \cmidrule{3-9} 
  
 \multicolumn{1}{l}{\textbf{Zero-Shot}} \\
\multicolumn{1}{l}{CLIP\cite{radford2021learning}}  & ICML'21 &69.4  &45.8  &64.3  &13.7  &82.7  &62.3  &56.4  \\
    \cmidrule{1-1} \cmidrule{2-2} \cmidrule{3-9} 
\multicolumn{1}{l}{\textbf{Non-CLIP Based}} \\
\multicolumn{1}{l}{ICON\cite{yue2023icon}}       & NeurIPS'23   & 76.2 & 33.6 & \textbf{76.1} & 12.1 & 84.1 & 63.9 & 57.7\\
\multicolumn{1}{l}{CSR\cite{zhou2024csr}}       &  AAAI'24   & 80.1 & 36.2 & 68.4 & \textbf{38.0} & 80.4 & 68.0 & 61.9\\
    \cmidrule{1-1} \cmidrule{2-2} \cmidrule{3-9} 
 \multicolumn{1}{l}{\textbf{CLIP Based}} \\

\multicolumn{1}{l}{AD-CLIP\cite{singha2023adclip}}    & ICCV'23    & 73.7 & 11.5 & 66.4 & 11.5 & 84.8 & 66.3 & 52.4 \\
\multicolumn{1}{l}{MPA\cite{chen2023mpa}}        & NeurIPS'23 & 75.4 & 49.2 & 69.4 & 15.5 & 83.7 & 65.3 & 59.8 \\
\multicolumn{1}{l}{DAPL\cite{ge2023dapl}}       & TNNLS'23    & 75.2 & 49.4 & 68.8 & 0.3  & 83.7 & 67.6 & 57.5 \\
\multicolumn{1}{l}{PDA\cite{bai2024pda}}        &  AAAI'24   & 76.1 & 50.0 & 70.6 & 17.0 & 84.1 & 67.1 & 60.8 \\

\multicolumn{1}{l}{PGA\cite{phan2024pga}}        & NeurIPS'24    & 75.1 & 51.1 & 69.4 & 14.5 & 84.5 & 67.4 & 60.3 \\
\multicolumn{1}{l}{DAMP\cite{du2024damp}}       &  CVPR'24  & 75.7 & 54.1 & 71.7 & 20.1 & 84.9 & 68.2 & 62.5 \\
    \cmidrule{1-1} \cmidrule{2-2} \cmidrule{3-9}

    \multicolumn{1}{l}{\textbf{\method} - Prompt}  & - & 78.2    & \textbf{55.1}    & 71.4    &17.9   & \textbf{86.1}  &70.0  & 63.1 \\
\multicolumn{1}{l}{\textbf{\method} - LoRA}  & - & 80.2     &54.0     & 72.6     &18.8   &85.7    &71.7  & 63.8 \\
\multicolumn{1}{l}{\textbf{\method} - Adapter}  & - & \textbf{81.0}     &52.8     & 73.2     &19.3 &85.7   &\textbf{72.4}    & \textbf{64.1} \\
\bottomrule
\end{tabular*}}
\vspace{3ex}
\label{tab:domainnet_performance}
\end{center}
\end{table*}

\subsection{Implementation Details}
We adopt CLIP ViT-B/16 as the backbone architecture for all methods. The textual prompts, visual PEFT modules, and autoencoder components are optimized using the Adam optimizer and CosineAnnealing scheduler. For the autoencoder, the projection layers are configured as follows: $W_2$ and $W_3$ generate intermediate and final embeddings of dimensions $e_2 = 384$ and $e_3 = 512$, while the input projection layer $W_1$ produces an embedding of dimension $e_1$, which is 150 for ImageCLEF and Office-Home, and 300 for DomainNet. 

The confidence threshold $\beta$ for the refine-and-rehearse phase is fixed at 0.8 for all datasets. The pseudo-label confidence threshold $\tau$ used during initial prompt learning is dataset-specific. We choose $\tau=0.4$ for DomainNet, and $\tau=0.6$ for ImageCLEF and Office-Home. For initial pseudo-label generation, we adopt an average-confidence voting strategy for ImageCLEF and Office-Home, and majority voting for DomainNet due to its larger number of domains. The number of curriculum clusters is set to $T=3$ for all datasets. Additional hyperparameter settings are detailed in the subsequent sections.

\subsection{Comparison to State-of-the-Art}

We present a comprehensive evaluation of our proposed method, \method, on three standard domain adaptation benchmarks. The performance on ImageCLEF and Office-Home is detailed in Table~\ref{tab:oh_clef_performance}, while the results for the more challenging DomainNet are in Table~\ref{tab:domainnet_performance}. For all experiments, we report the performance of \method~integrated with three distinct PEFT modules: Visual Prompts, LoRA, and Adapters, as discussed in Section~\ref{sec:architectural design}.

On the ImageCLEF dataset, \method
~with LoRA establishes a new state-of-the-art, achieving an average accuracy of 94.3\%. This represents a significant 1.5\% improvement over the next-best method, ICON. Specifically, our approach demonstrates consistently superior performance across all domains, with accuracies of 98.8\% on domain C, 99.0\% on domain I, and 85.0\% on P. Given the small scale of ImageCLEF and the inherent strength of the CLIP backbone, these gains are particularly noteworthy. The \method-Prompt variant also surpasses existing methods, securing an average accuracy of 92.9\%. In contrast, the \method-Adapter shows a marked performance degradation on this dataset, with its average accuracy dropping to 89.4\%, suggesting potential optimization challenges on smaller-scale benchmarks.

Moving to the more complex Office-Home dataset, \method-LoRA continues its impressive performance, achieving an average accuracy of 91.8\%. This result outperforms all prior approaches by a substantial margin of at least 2.6\%. The method achieves the highest accuracy across all four domains, with 93.6\% on Art, 83.2\% on Clipart, 94.8\% on Product, and 95.6\% on Real World. These consistent gains underscore the robustness of our progressive alignment strategy in learning domain-invariant features across diverse visual styles. Following a similar trend to ImageCLEF, \method-Prompt also delivers strong results, outperforming the next-best method, ICON, by 0.3\%. Notably, while \method-Adapter still does not achieve the top performance, its accuracy of 88.9\% is competitive, falling short only of ICON but outperforming all other baselines.

Finally, on the most challenging benchmark, DomainNet, we observe a notable shift in the efficacy of the PEFT modules. Here, \method-Adapter achieves the highest overall accuracy at 64.1\%, surpassing all previous state-of-the-art methods. It improves upon DAMP, a strong CLIP-based baseline, by 1.6\% and significantly outperforms zero-shot CLIP by 7.1\%. This superior performance suggests that the Adapter module's architectural complexity may require larger data for effective optimization. This finding is consistent with its weaker performance on the smaller ImageCLEF dataset and better performance on the mid-scale Office-Home. While not the top performer in this instance, \method-LoRA still demonstrates strong results with an accuracy of 63.8\%, outperforming DAMP by 1.3\%. Furthermore, \method-Prompt also surpasses other state-of-the-art methods with an average accuracy of 63.1\%. The fact that all three PEFT variants of our method outperform existing approaches on this difficult benchmark highlights the remarkable effectiveness of our proposed method in handling severe domain shifts, noisy pseudo-labels, and the inherent complexity of large-scale, multi-source adaptation.

\begin{figure*}[!t]
\centering
\subfloat[Visual Prompt.]{%
    \includegraphics[width=0.32\linewidth]{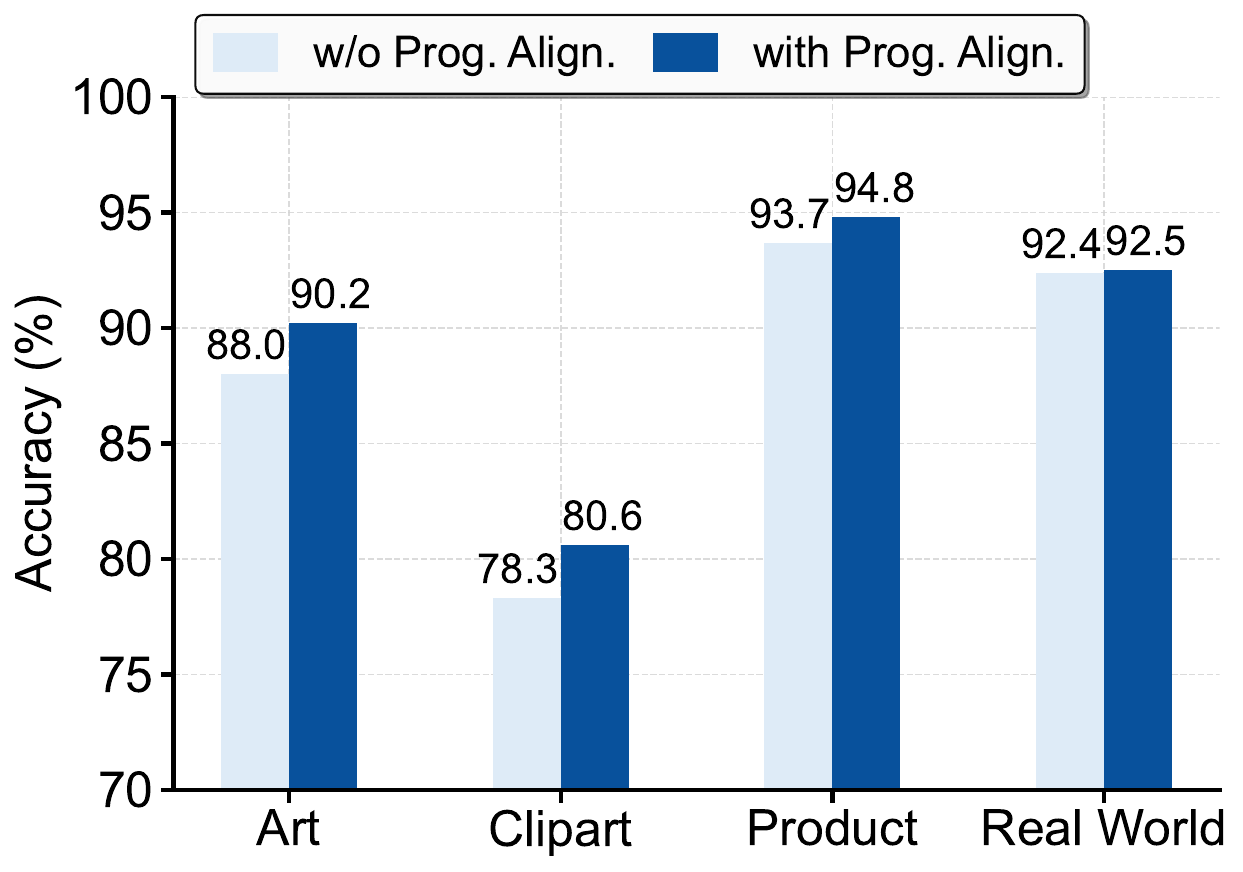}%
    \label{fig:prompt_strategy_performance}
}
\hfill
\subfloat[LoRA.]{%
    \includegraphics[width=0.32\linewidth]{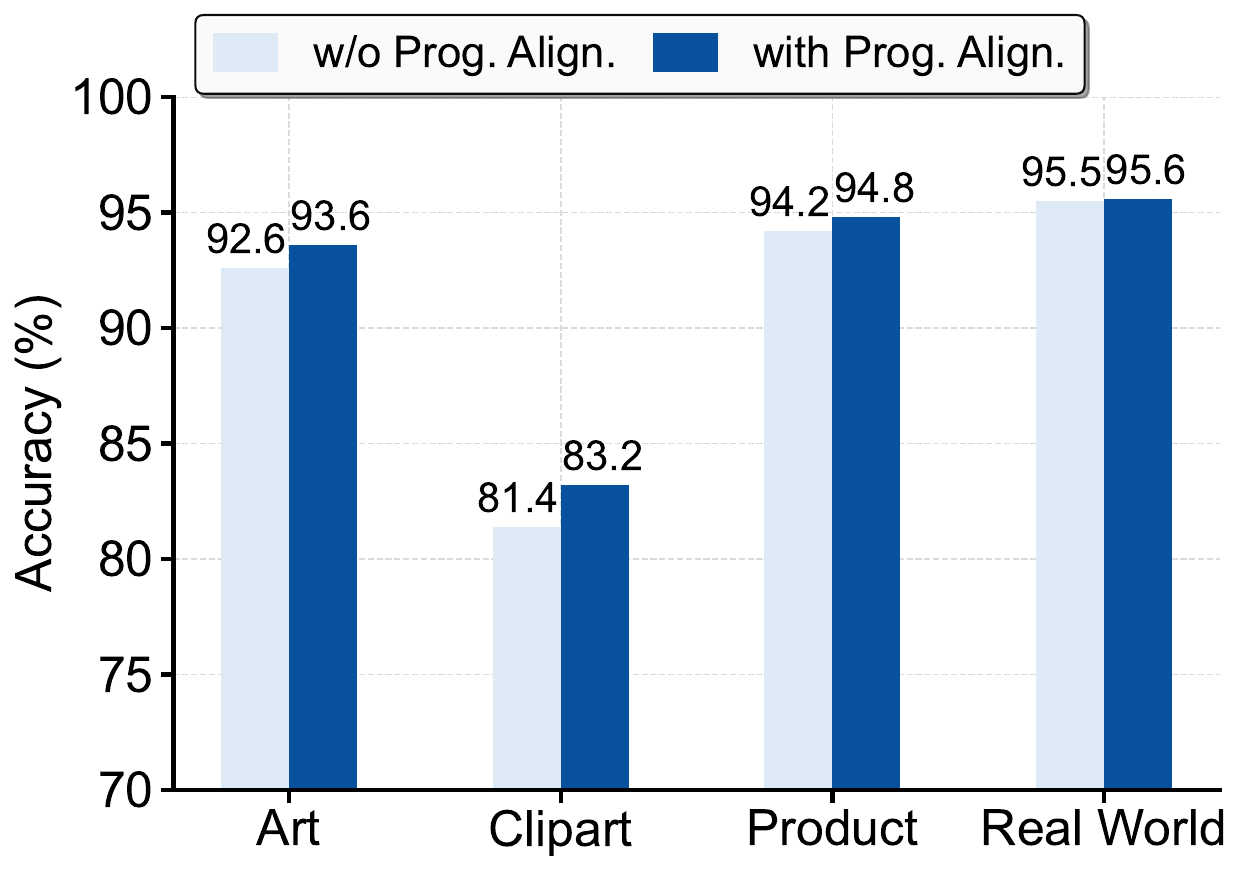}%
    \label{fig:lora_strategy_performance}
}
\hfill
\subfloat[Adapter.]{%
    \includegraphics[width=0.32\linewidth]{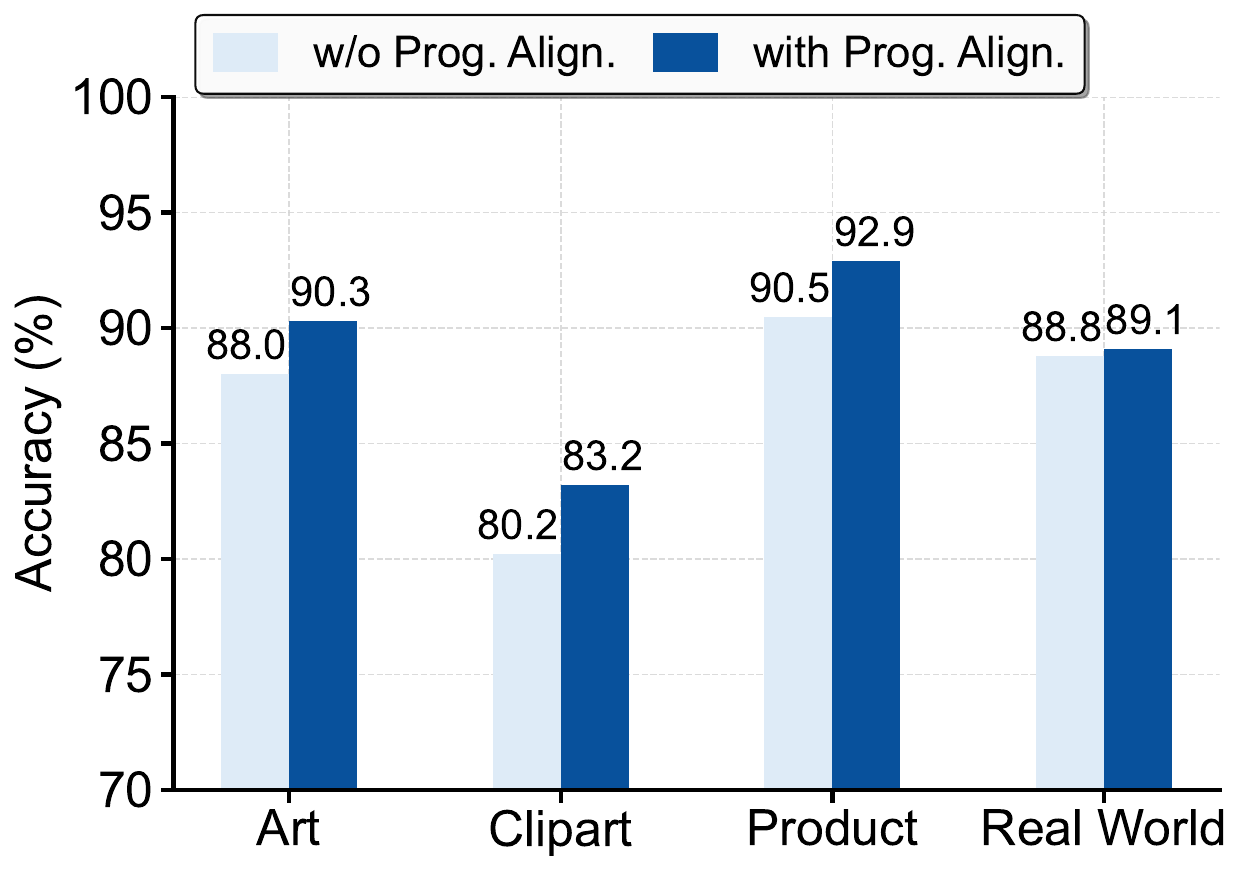}%
    \label{fig:adapter_strategy_performance}
}
\caption{Empirical analysis on the impact of the progressive alignment strategy using Visual Prompt, LoRA, and Adapter.}
\label{fig:combined_strategy_peft}
\end{figure*}

\begin{table*}[t]
\caption{Ablation Study on Cluster Center.}
\centering
\setlength{\tabcolsep}{3.5pt}
\resizebox{0.95\linewidth}{!}{%
\begin{tabular}{c|ccccc|ccccc|ccccc}
\toprule
 & \multicolumn{5}{c|}{\textbf{Visual Prompt}} 
 & \multicolumn{5}{c|}{\textbf{LoRA}} 
 & \multicolumn{5}{c}{\textbf{Adapter}} \\ 
 \addlinespace[0.5ex]
\textbf{Cluster Number} & \textbf{$\rightarrow$ Ar} & \textbf{$\rightarrow$ Cl} & \textbf{$\rightarrow$ Pr} & \textbf{$\rightarrow$ Rw} & \textbf{Avg} 
& \textbf{$\rightarrow$ Ar} & \textbf{$\rightarrow$ Cl} & \textbf{$\rightarrow$ Pr} & \textbf{$\rightarrow$ Rw} & \textbf{Avg}
& \textbf{$\rightarrow$ Ar} & \textbf{$\rightarrow$ Cl} & \textbf{$\rightarrow$ Pr} & \textbf{$\rightarrow$ Rw} & \textbf{Avg} \\
\cmidrule(lr){1-1} \cmidrule(lr){2-6} \cmidrule(lr){7-11} \cmidrule(lr){12-16}

$T = 2$ & 89.3 & 79.8 & 94.0 & 93.2 & 88.9 
& 93.4 & 83.0 & 94.4 & 94.8 & 91.4
& 90.6 & 82.4 & 92.2 & 89.5 & 88.7 \\ 

\addlinespace[0.5ex]

$T = 5$ & 90.1 & 79.0 & 94.8 & 92.2 & 89.0
& 92.9 & 83.5 & 94.2 & 95.4 & 91.5
& 90.4 & 80.6 & 92.2 & 88.9 & 88.0 \\

\addlinespace[0.5ex]

$T = 10$ & 89.4 & 78.7 & 94.0 & 92.2 & 88.6
& 92.0 & 82.7 & 94.1 & 94.7 & 90.9
& 88.3 & 81.7 & 93.0 & 89.4 & 88.1 \\

% \addlinespace[0.5ex]
\cmidrule(lr){1-1} \cmidrule(lr){2-6} \cmidrule(lr){7-11} \cmidrule(lr){12-16}
\rowcolor{gray!15}
$T = 3$ & 90.2 & 80.6 & 94.8 & 92.5 & \textbf{89.5}
& 93.4 & 83.6 & 94.8 & 95.6 & \textbf{91.9}
& 90.3 & 83.2 & 92.9 & 89.5 & \textbf{88.9} \\

\bottomrule
\end{tabular}%
}
\label{tab:cluster_ablation}
\end{table*}

\begin{figure}[t]
\centering
\subfloat[Without Progressive Alignment]{\includegraphics[width=0.48\linewidth]{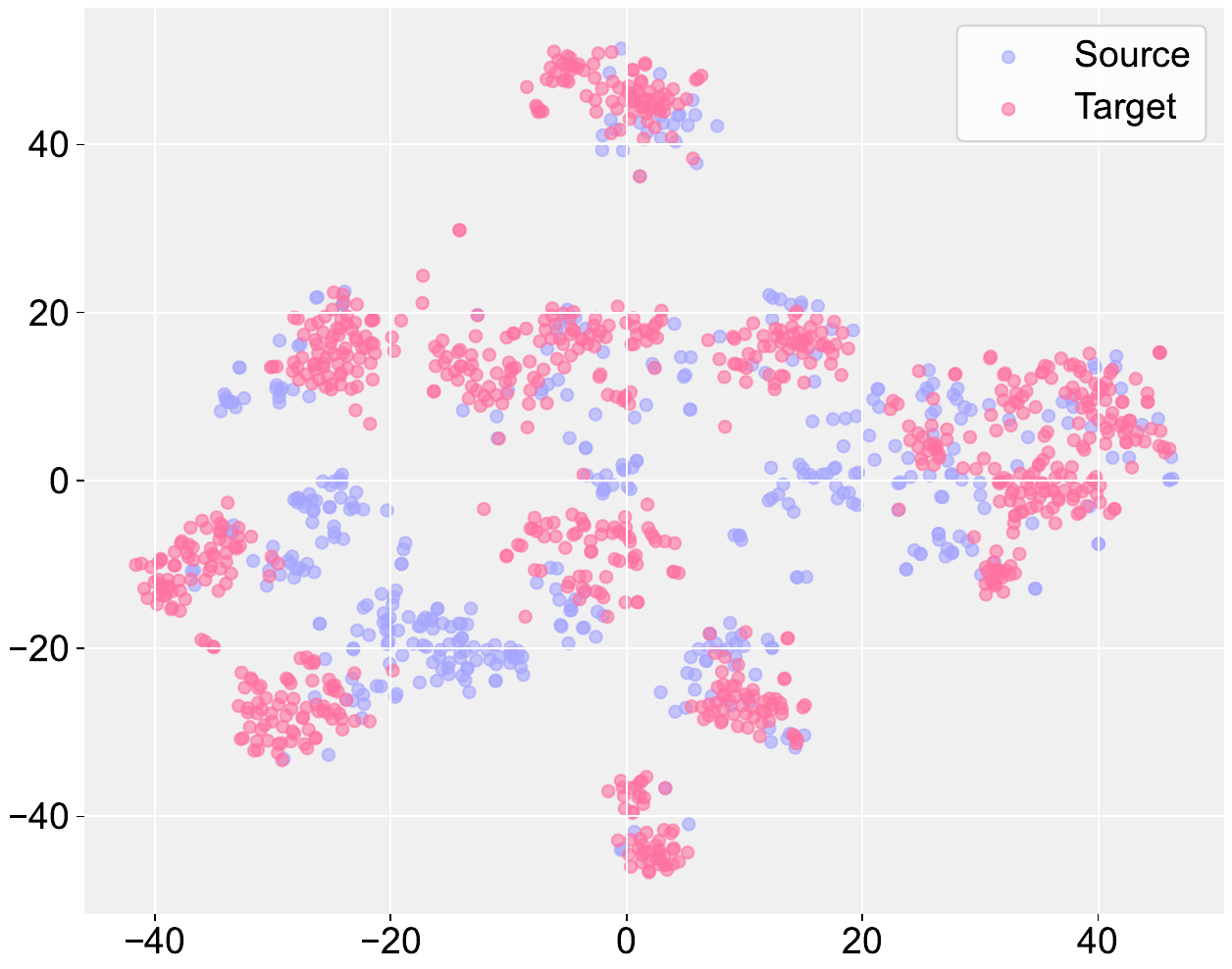}%
\label{fig_without_visual}}
\hfill
\subfloat[With Progressive Alignment]{\includegraphics[width=0.48\linewidth]{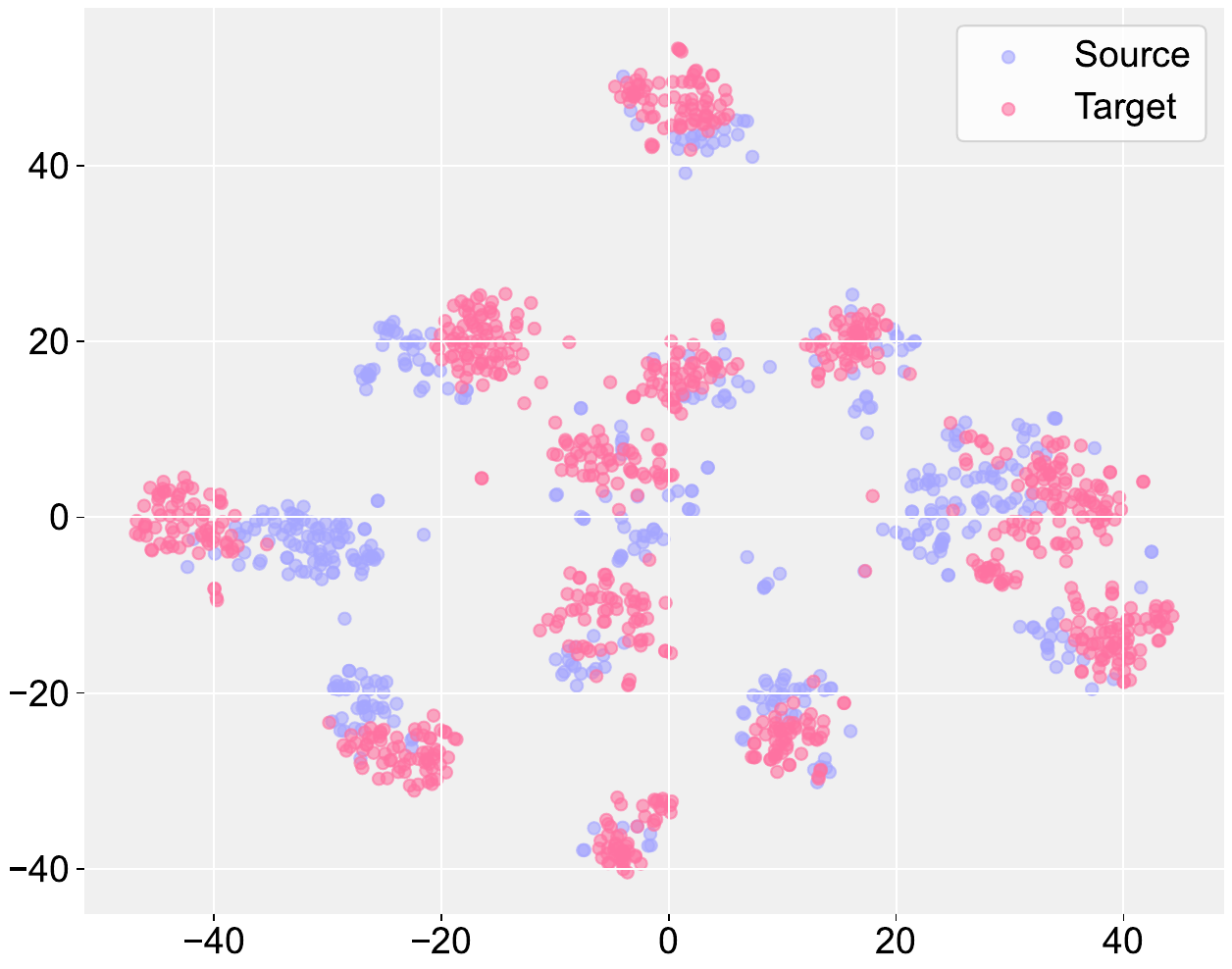}%
\label{fig_with_visual}}
\caption{t-SNE visualization of features extracted by the visual encoder using Adapter on the Office-Home dataset, with Art as the source domain and Real World as the target domain.}
\label{fig_sim}
\end{figure}

\subsection{Analysis on Progressive Alignment}

To rigorously evaluate the efficacy of our progressive alignment strategy, we conduct a detailed empirical analysis on the Office-Home dataset. This analysis first systematically compares the performance of all three PEFT modules with and without the progressive alignment component. The results are shown in Figure~\ref{fig:combined_strategy_peft}. As observed, incorporating progressive alignment consistently leads to performance improvements across all PEFT modules, demonstrating the broad applicability of the proposed strategy.

Specifically, we observe average performance gains of 1.4\% for Visual Prompt, 0.9\% for LoRA, and 2.1\% for Adapter, respectively. These consistent gains show the value of the proposed progressive alignment. By guiding the model through a structured, easy-to-hard curriculum, our strategy effectively mitigates the impact of noisy pseudo-labels, particularly during the critical early stages of training, thereby enhancing the model's robustness.

Through a closer inspection of the domain-wise performance, we further reveal insightful patterns into the mechanism of our approach. For the Real World domain, which shares a closer visual resemblance to the source domains, the performance improvement is minimal, averaging only 0.2\%. In contrast, the Clipart domain, characterized by its abstract and stylized visual properties, exhibits a substantial average improvement of 2.4\%. This significant gain suggests that our progressive alignment strategy is particularly advantageous for target domains with a large distributional gap from the source, as it systematically bridges this divide by starting with more easy and transferable features before progressing to more hard and domain-specific ones.

To qualitatively illustrate the effect of our progressive alignment strategy, we visualize the feature space alignment between the source and target domains. Using the t-SNE algorithm, we plot the features extracted by the visual encoder for the Office-Home dataset, with the Art domain serving as the source and Real World as the target. As shown in Figure~\ref{fig_sim}, a baseline model without our strategy exhibits a clear separation between the source and target feature clusters. In contrast, after applying our proposed method, the features from both domains become closely aligned. This significant increase in alignment demonstrates that our strategy successfully bridges the domain gap, confirming its effectiveness in facilitating the learning of robust, domain-invariant representations.

\begin{figure*}[!t]
\centering
\subfloat[Visual Prompt.]{%
    \includegraphics[width=0.32\linewidth]{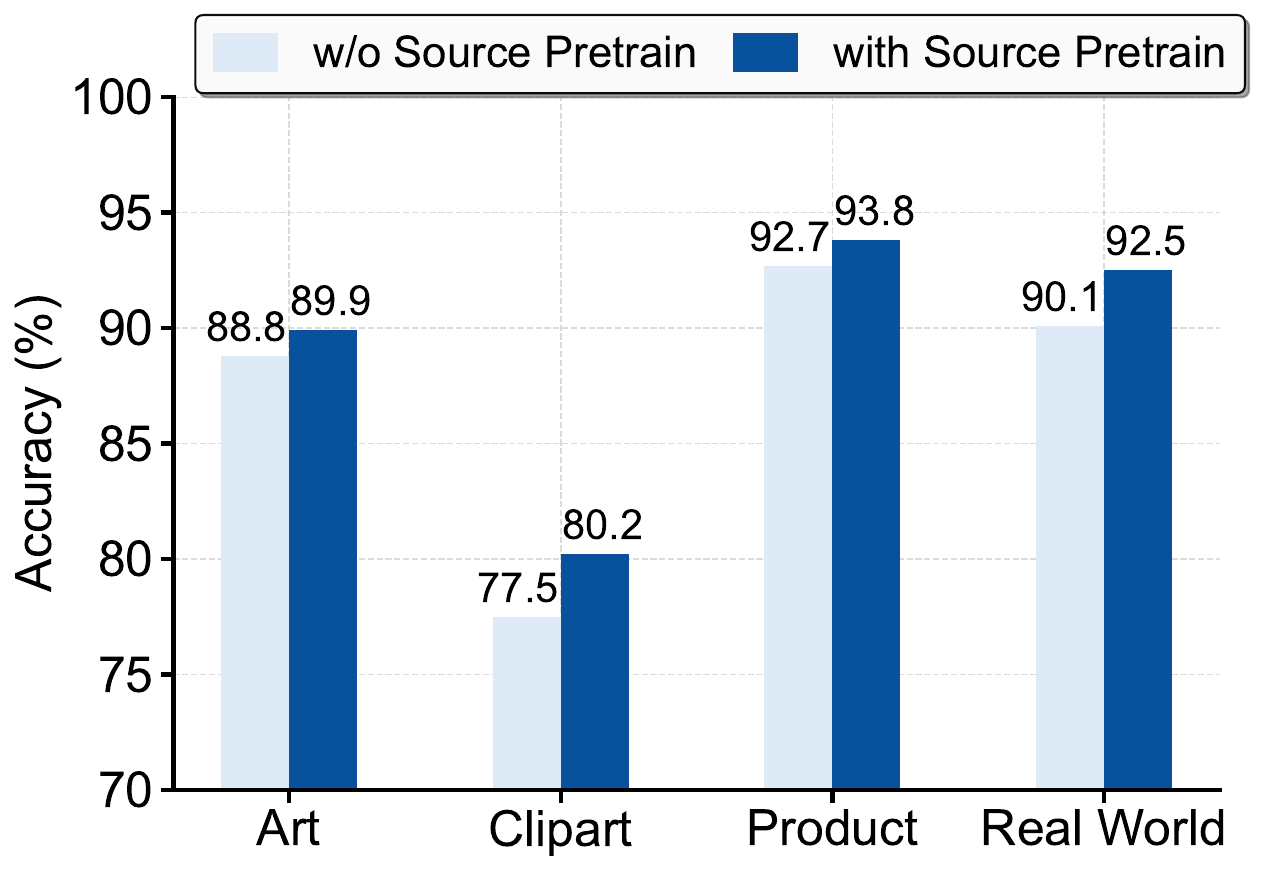}%
    \label{fig:prompt_source_pretrain}
}
\hfill
\subfloat[LoRA.]{%
    \includegraphics[width=0.32\linewidth]{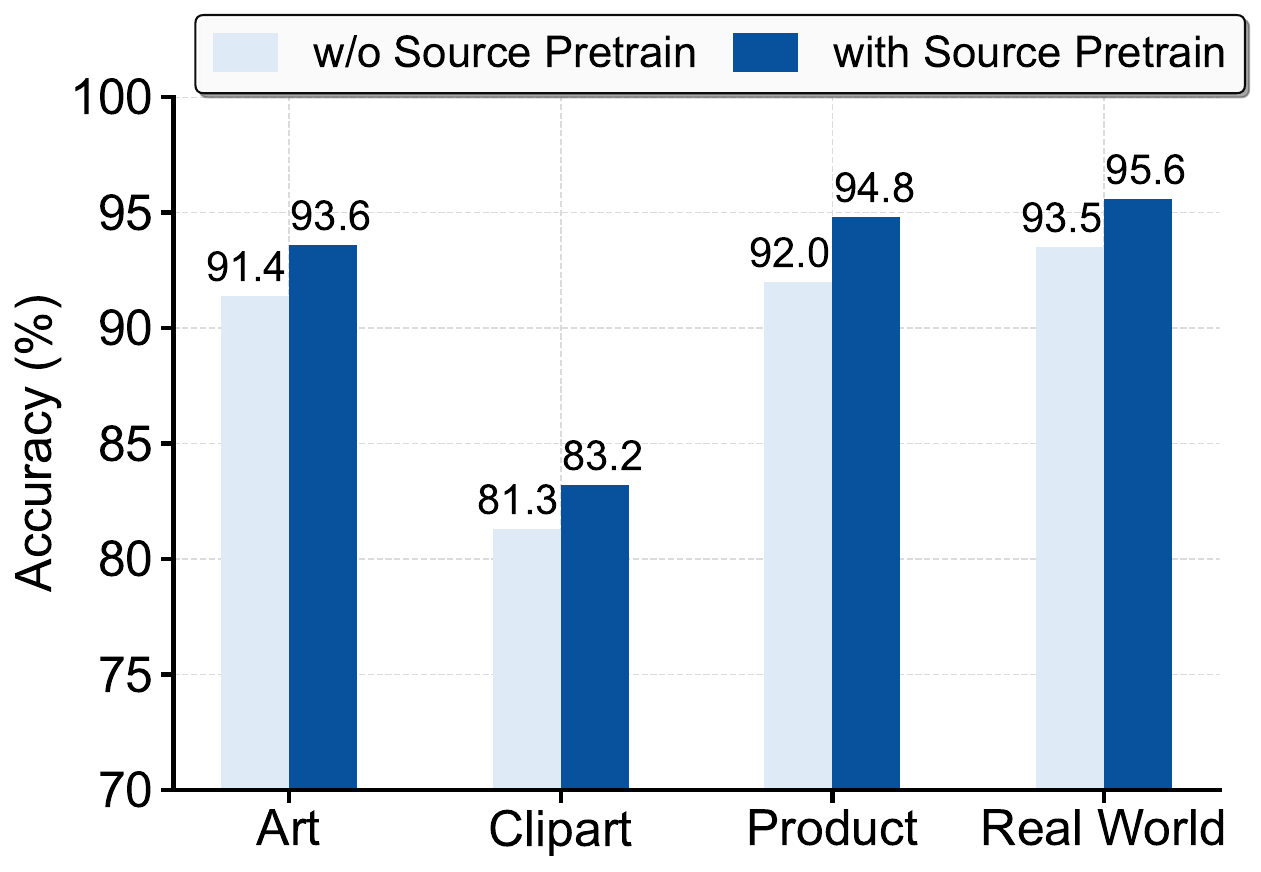}%
    \label{fig:lora_source_pretrain}
}
\hfill
\subfloat[Adapter.]{%
    \includegraphics[width=0.32\linewidth]{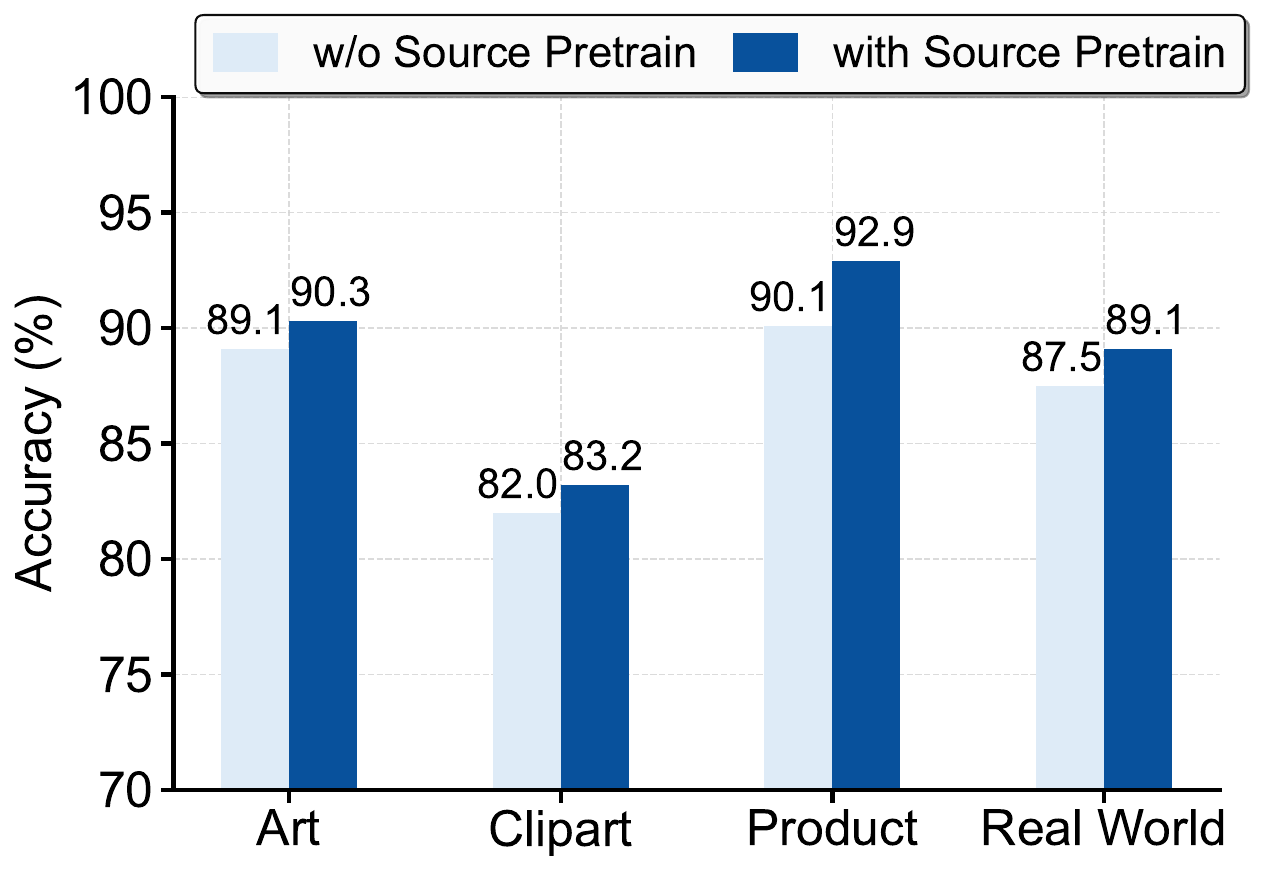}%
    \label{fig:adapter_source_pretrain}
}
\caption{Empirical analysis on the impact of the source pretrain strategy using Visual Prompt, LoRA, and Adapter.}
% \vspace{-2ex}
\label{fig:combined_source_pretrain}
\end{figure*}

\begin{figure*}[!t]
\centering
\subfloat[Ablation study on $\tau$]{%
    \includegraphics[width=0.24\linewidth]{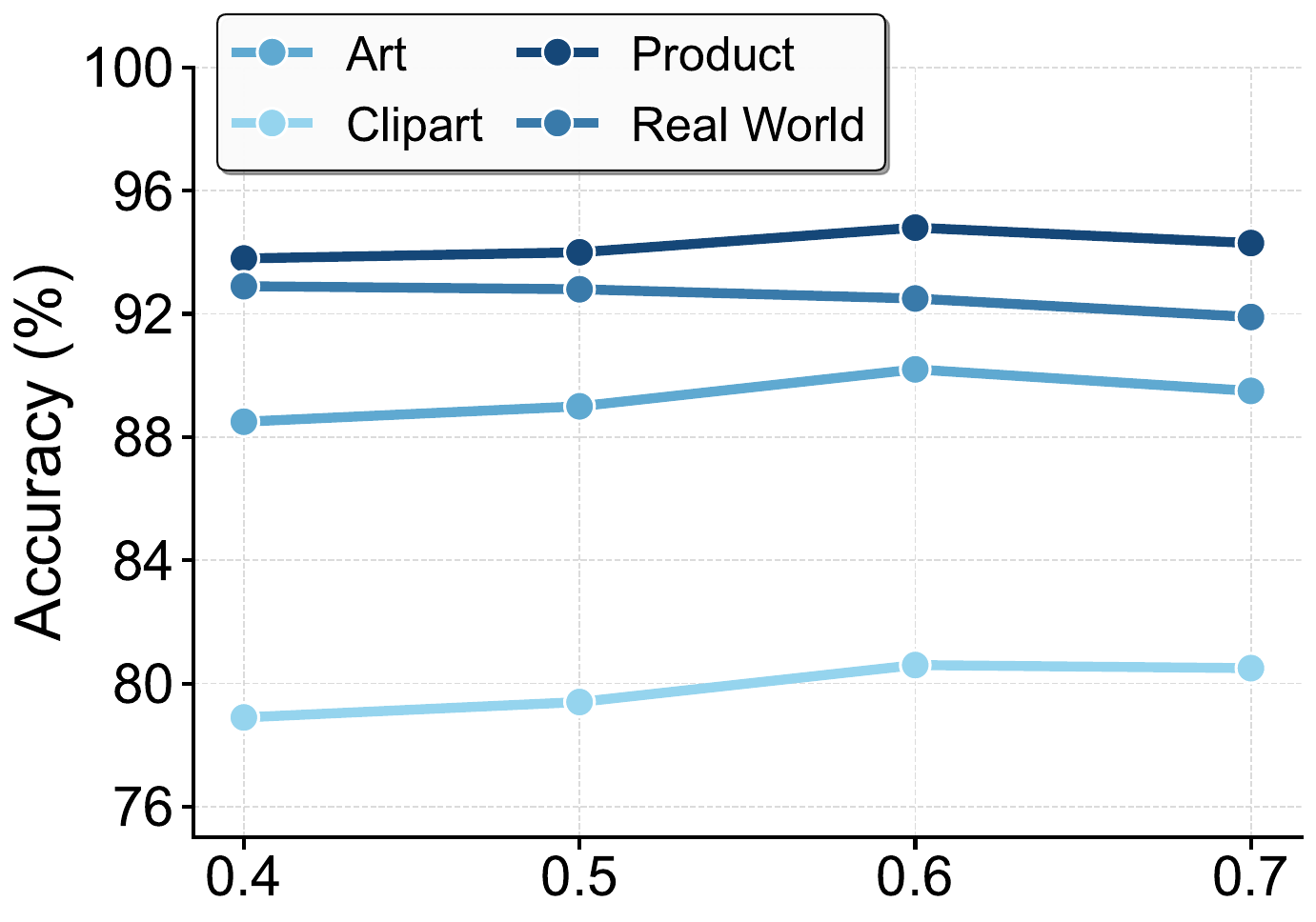}%
    \label{fig:prompt_tau}
}
\hfill
\subfloat[Ablation study on $\beta$]{%
    \includegraphics[width=0.24\linewidth]{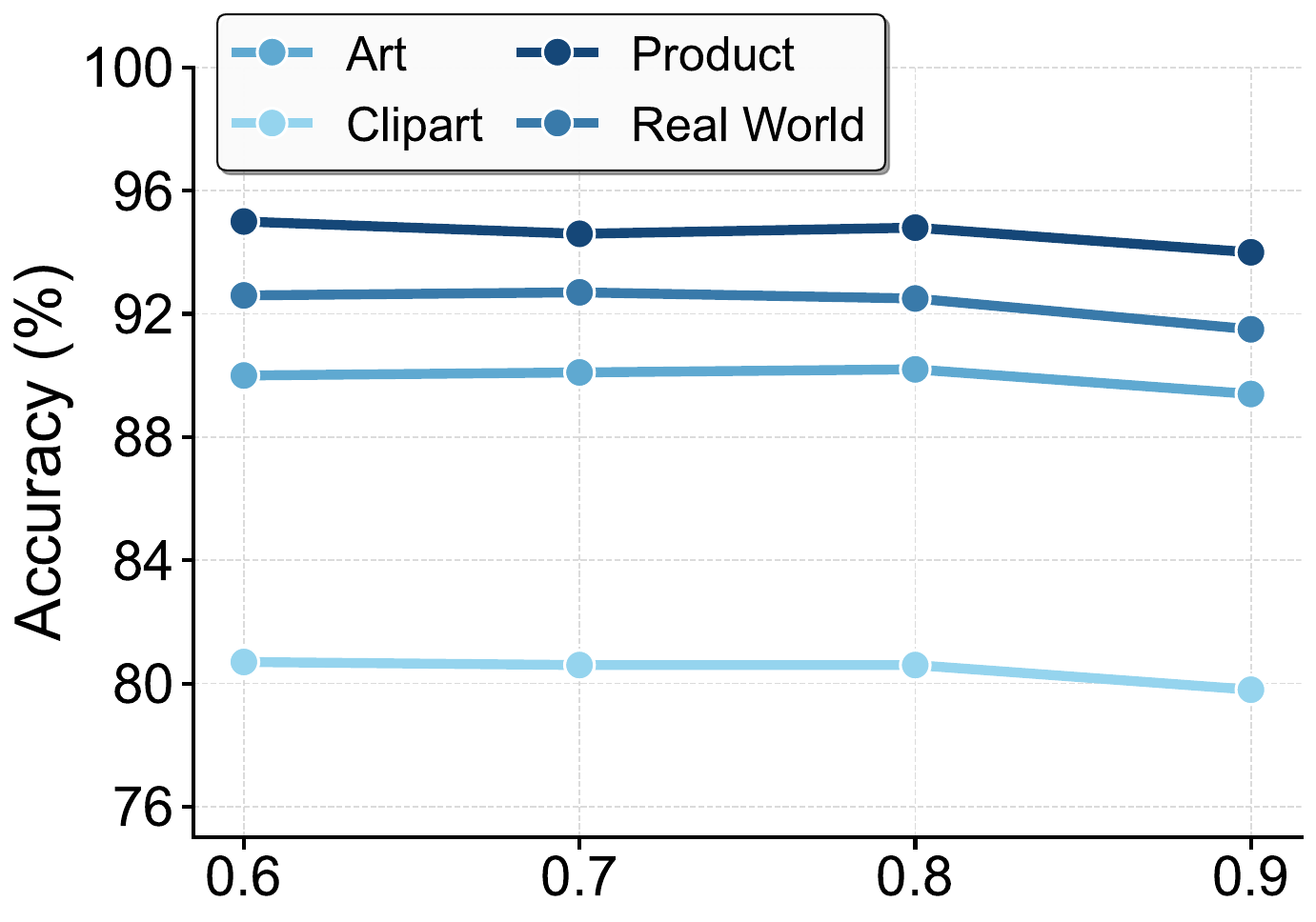}%
    \label{fig:prompt_beta}
}
\hfill
\subfloat[Ablation study on $M_1$ and $M_2$]{%
    \includegraphics[width=0.24\linewidth]{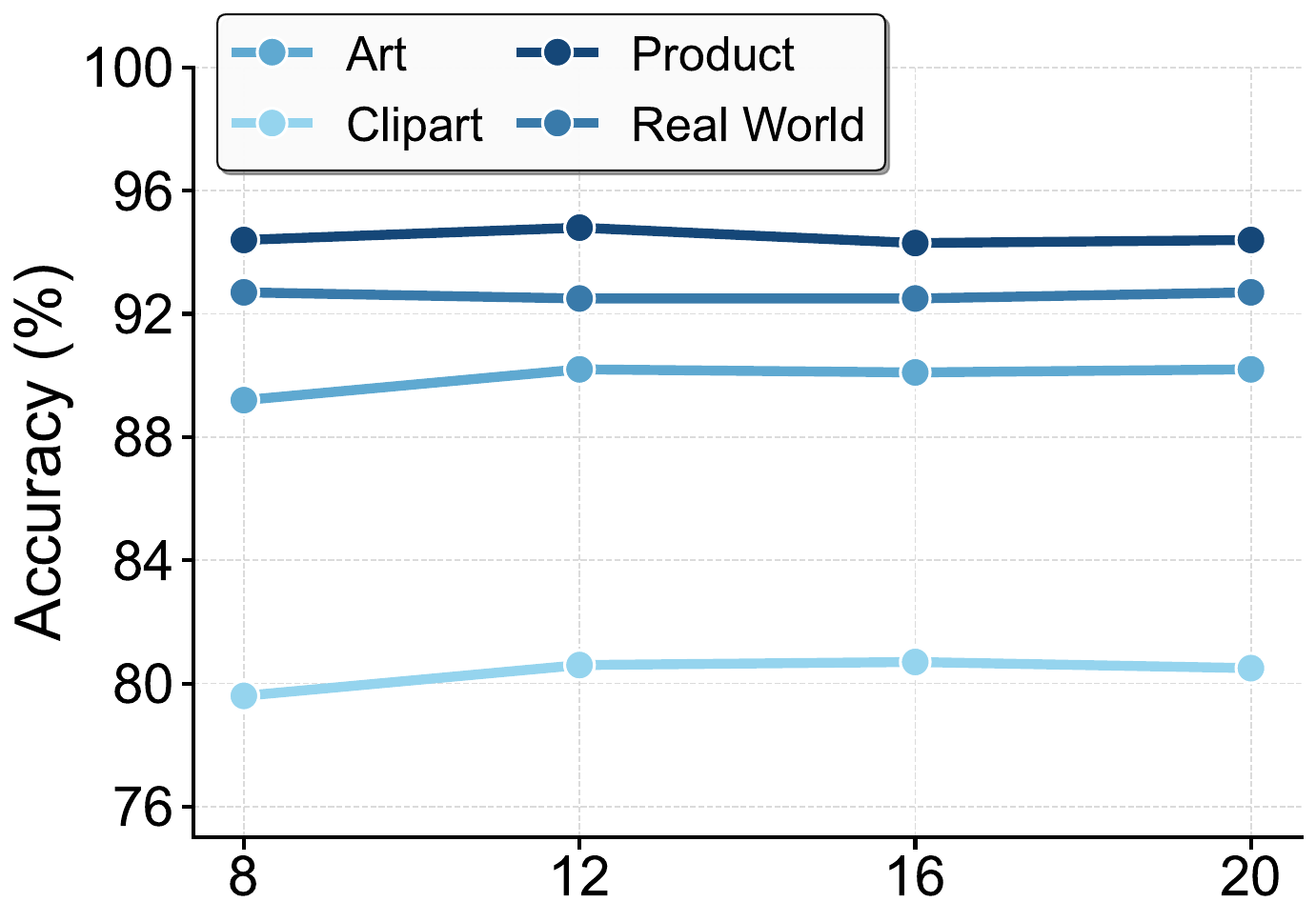}%
    \label{fig:prompt_m1_m2}
}
\hfill
\subfloat[Ablation study on $M_3$]{%
    \includegraphics[width=0.24\linewidth]{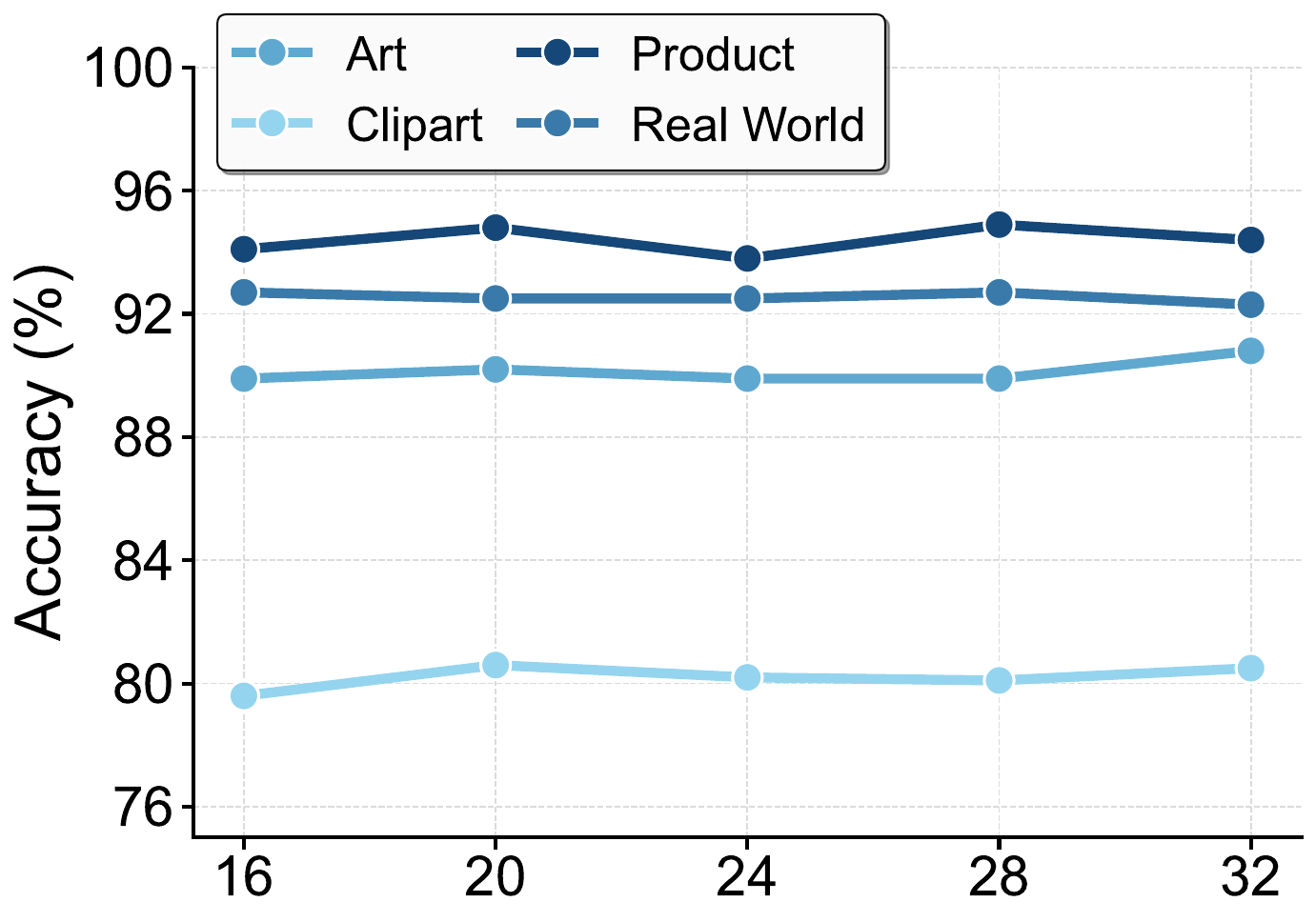}%
    \label{fig:prompt_m3}
}
\caption{\method~- Prompt ablation studies. (a) threshold parameter $\tau$ (b) threshold parameter $\beta$ (c) textual prompt length $M_1$ and $M_2$ (d) visual prompt length $M_3$.}
\label{fig:ablation_prompt}
% \vspace{-2ex}
\end{figure*}

\begin{figure*}[!t]
\centering
\subfloat[Ablation study on $\tau$]{%
    \includegraphics[width=0.24\linewidth]{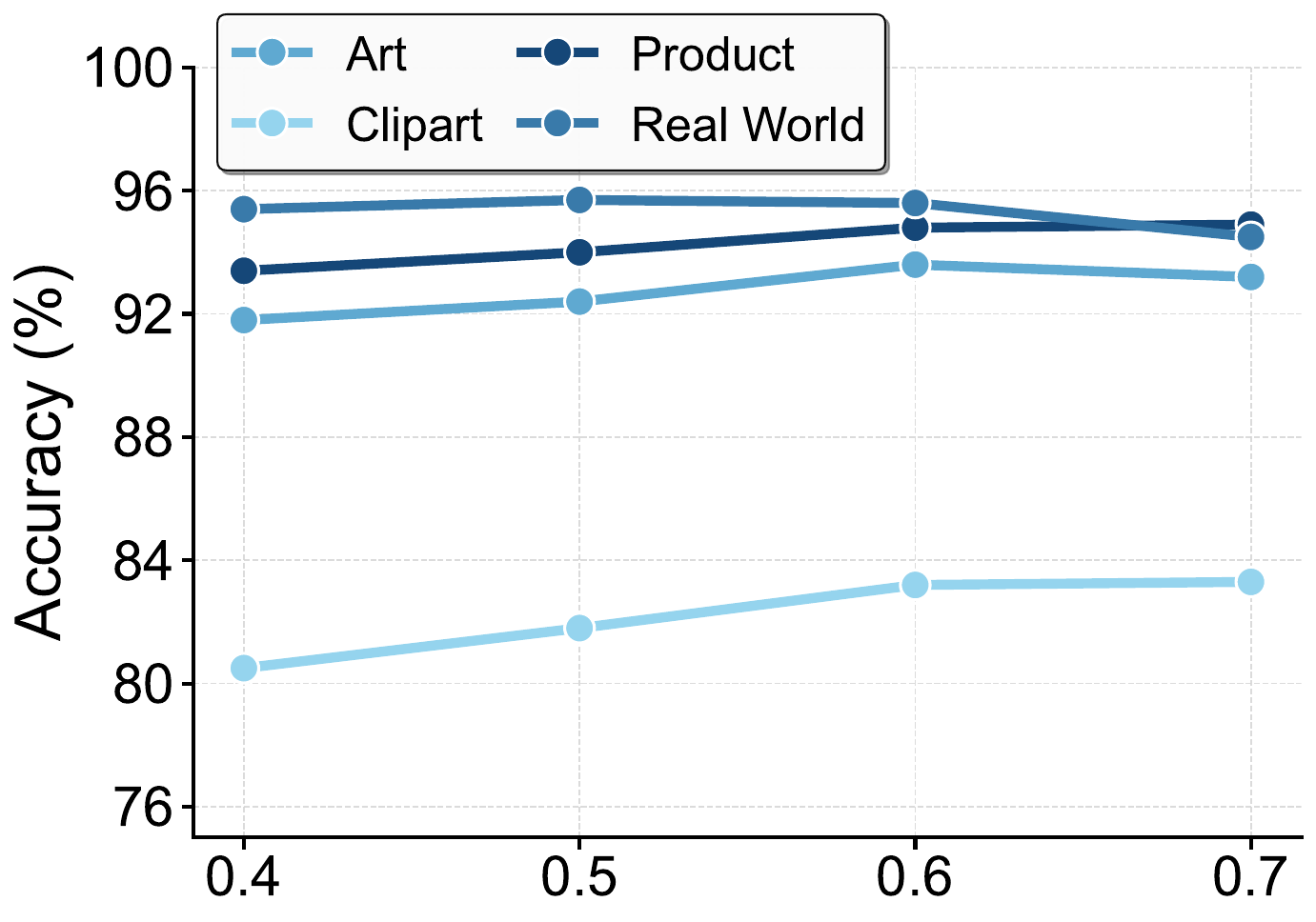}%
    \label{fig:lora_tau}
}
\hfill
\subfloat[Ablation study on $\beta$]{%
    \includegraphics[width=0.24\linewidth]{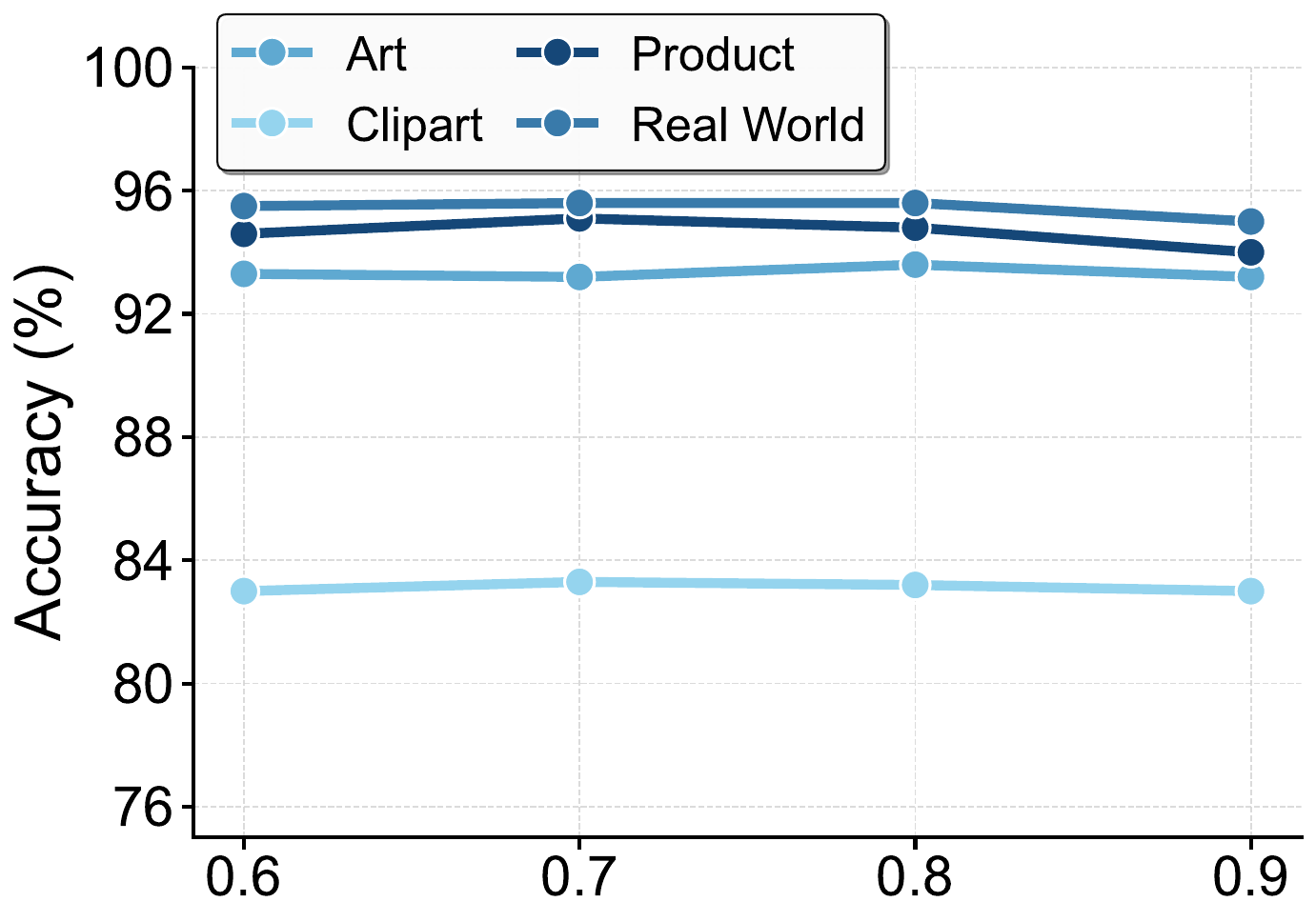}%
    \label{fig:lora_beta}
}
\hfill
\subfloat[Ablation study on $M_1$ and $M_2$]{%
    \includegraphics[width=0.24\linewidth]{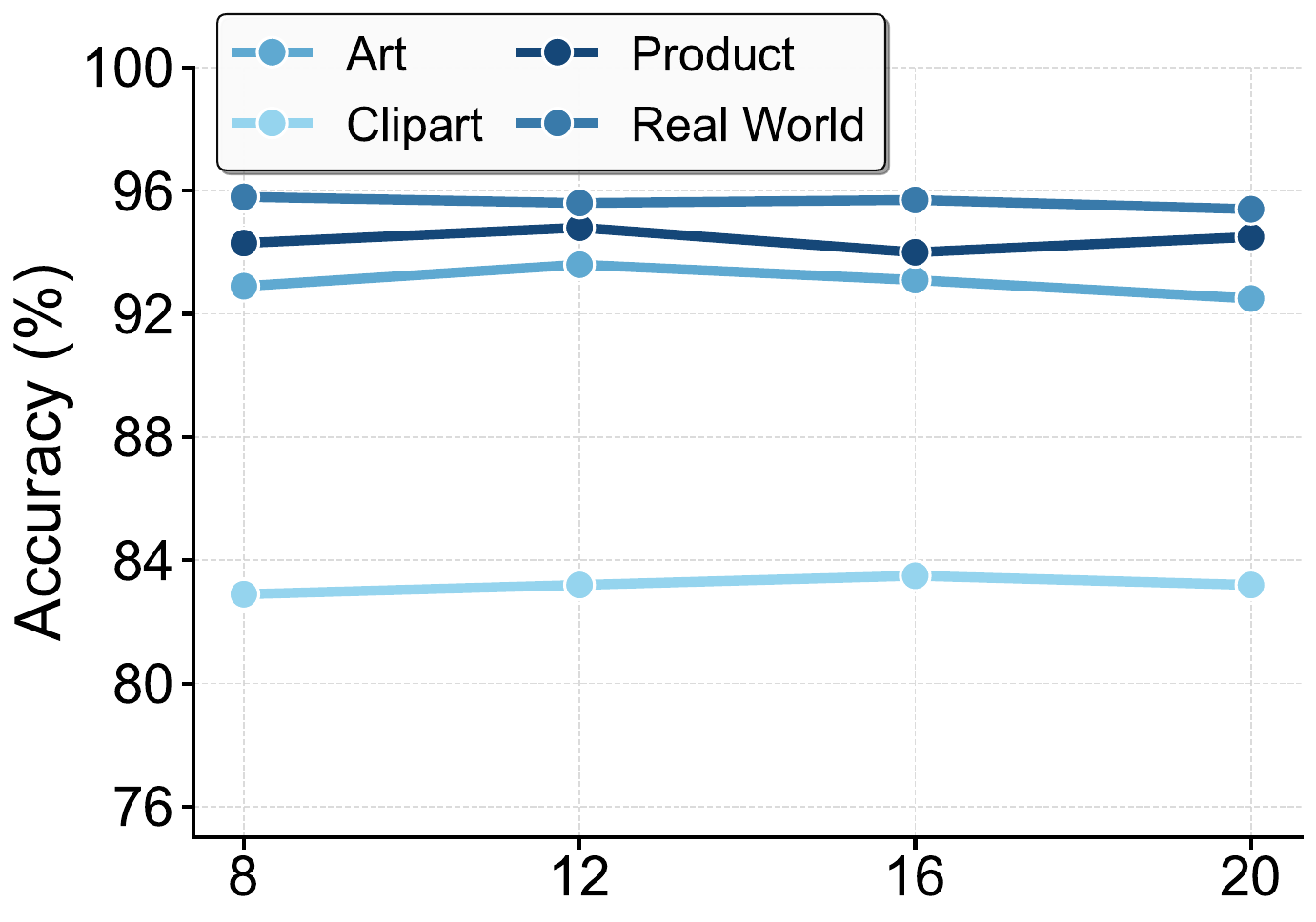}%
    \label{fig:lora_m1_m2}
}
\hfill
\subfloat[Ablation study on $r_1$]{%
    \includegraphics[width=0.24\linewidth]{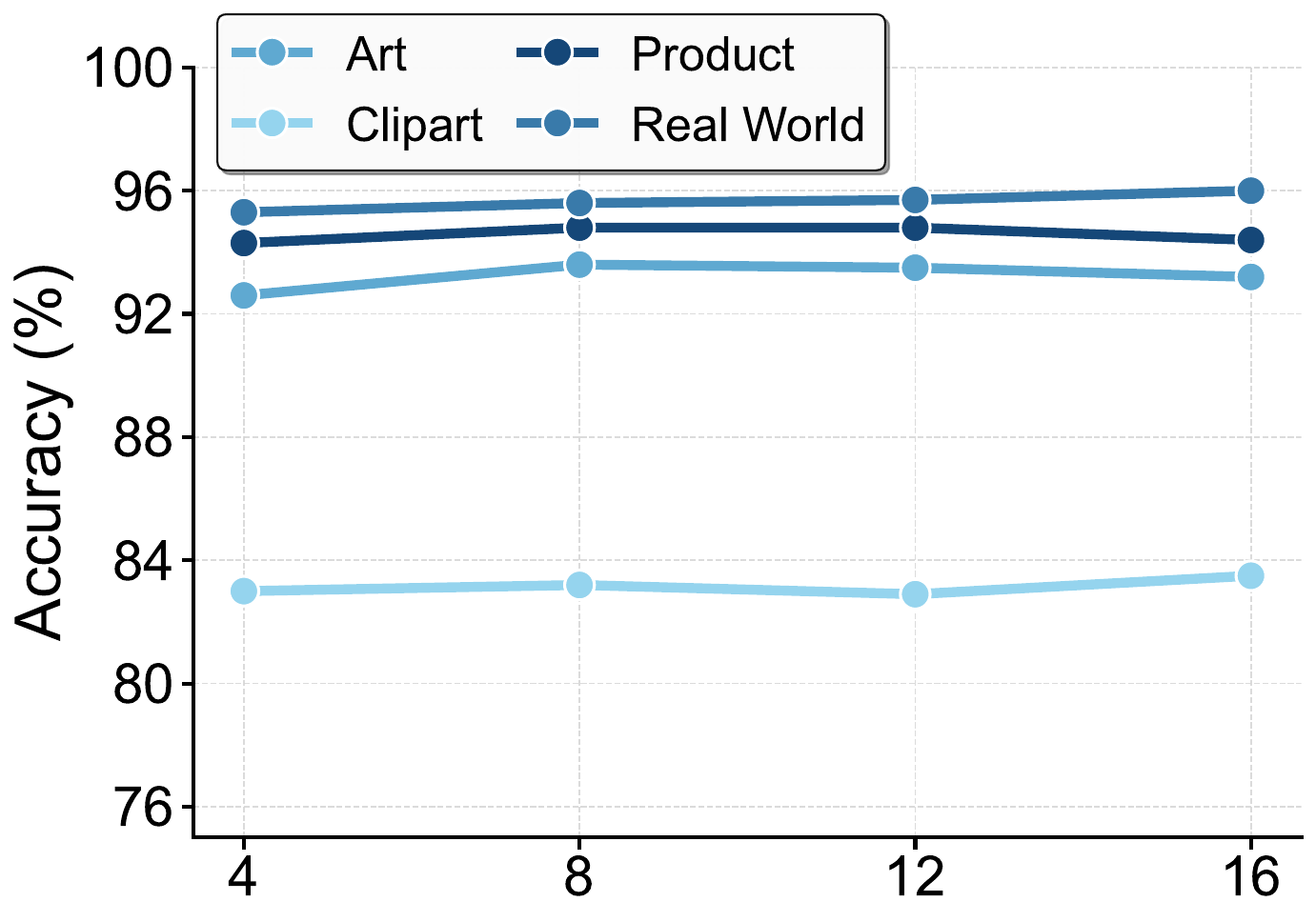}%
    \label{fig:lora_r}
}
\caption{\method~- LoRA ablation studies. (a) threshold parameter $\beta$ (b) threshold parameter $\tau$ (c) prompt length $M_1$ and $M_2$ (d) lora rank $r_1$.}
\label{fig:ablation_lora}
\end{figure*}

\begin{figure*}[!t]
\centering
\subfloat[Ablation study on $\tau$]{%
    \includegraphics[width=0.24\linewidth]{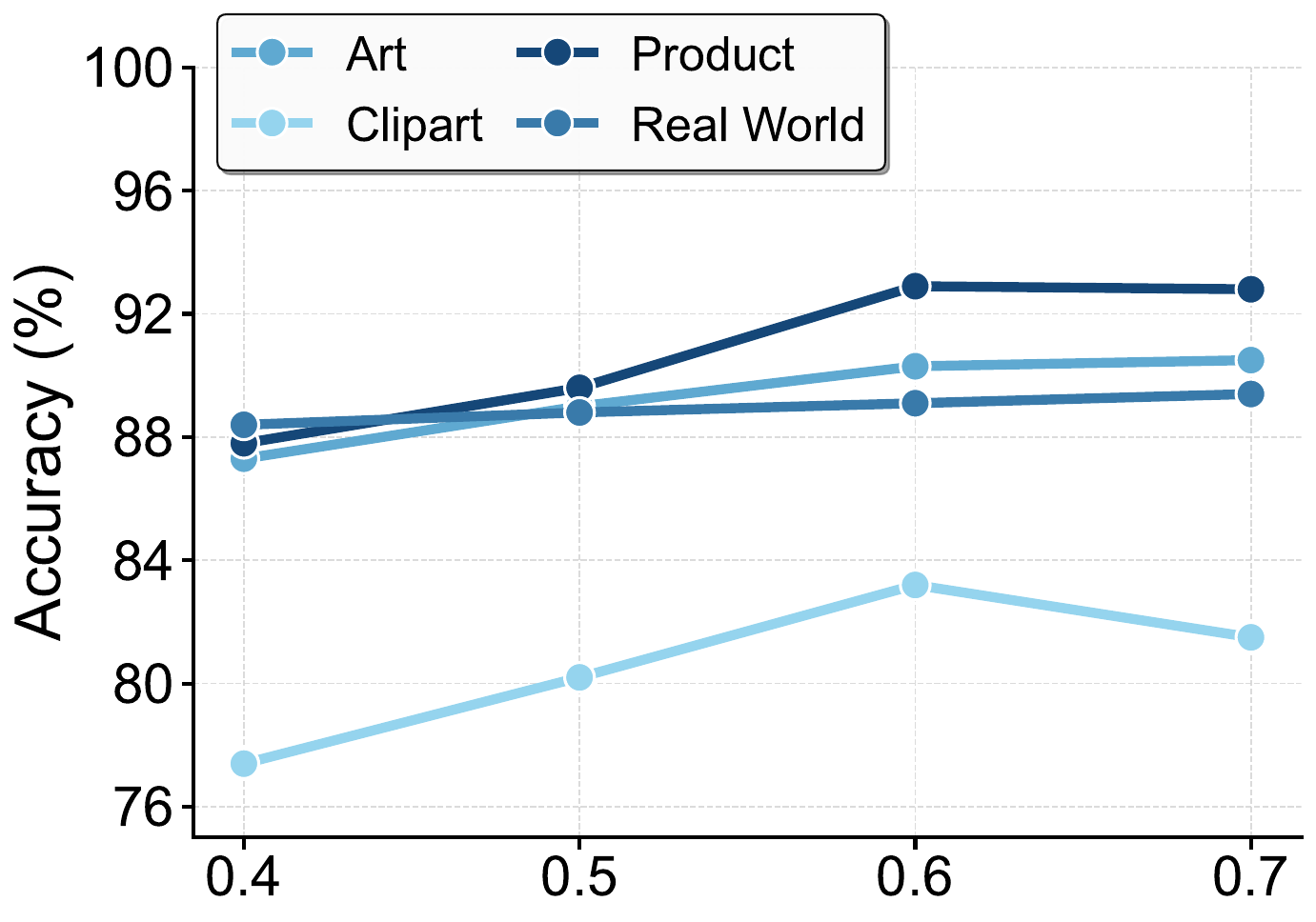}%
    \label{fig:adapter_tau}
}
\hfill
\subfloat[Ablation study on $\beta$]{%
    \includegraphics[width=0.24\linewidth]{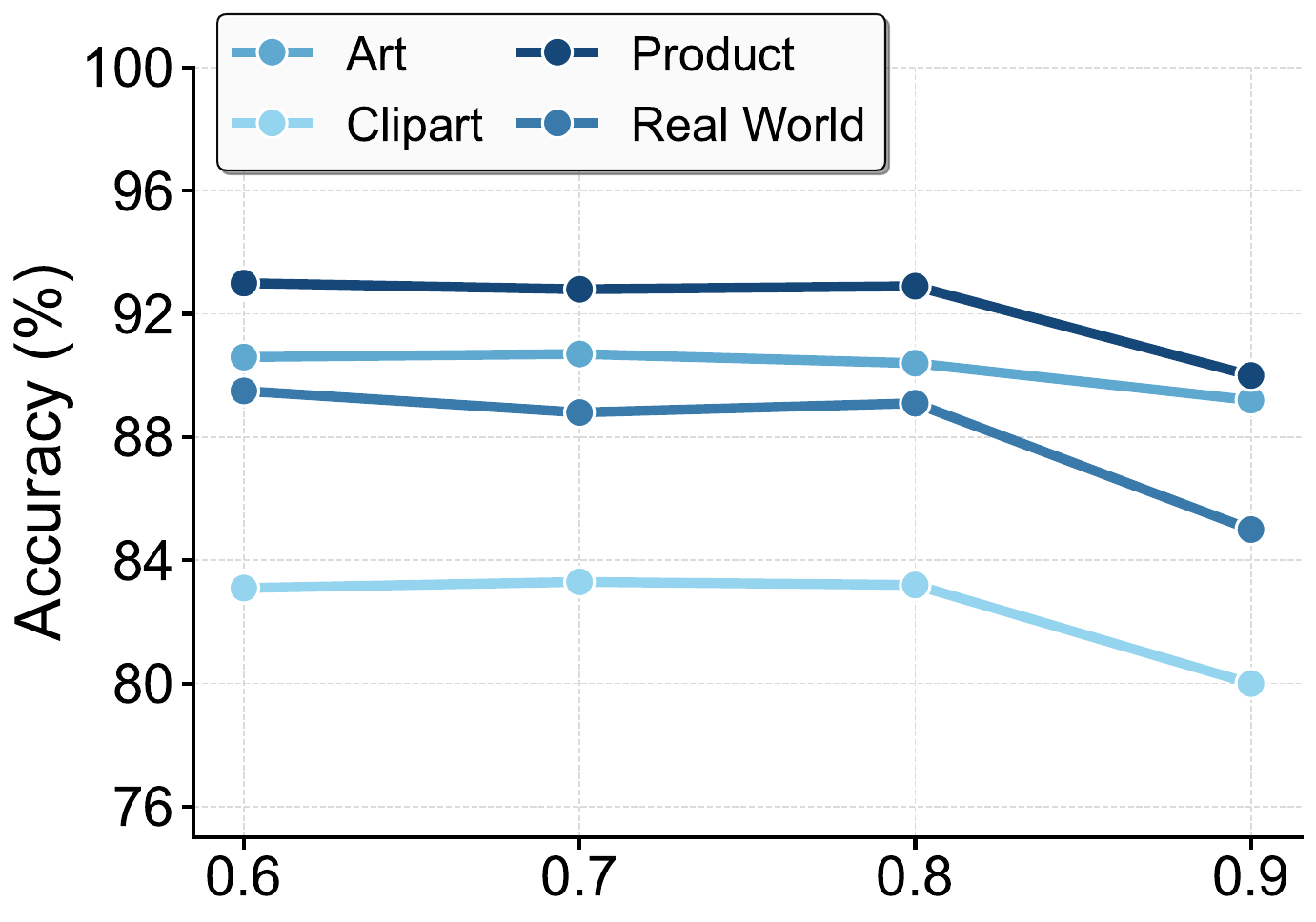}%
    \label{fig:adapter_beta}
}
\hfill
\subfloat[Ablation study on $M_1$ and $M_2$]{%
    \includegraphics[width=0.24\linewidth]{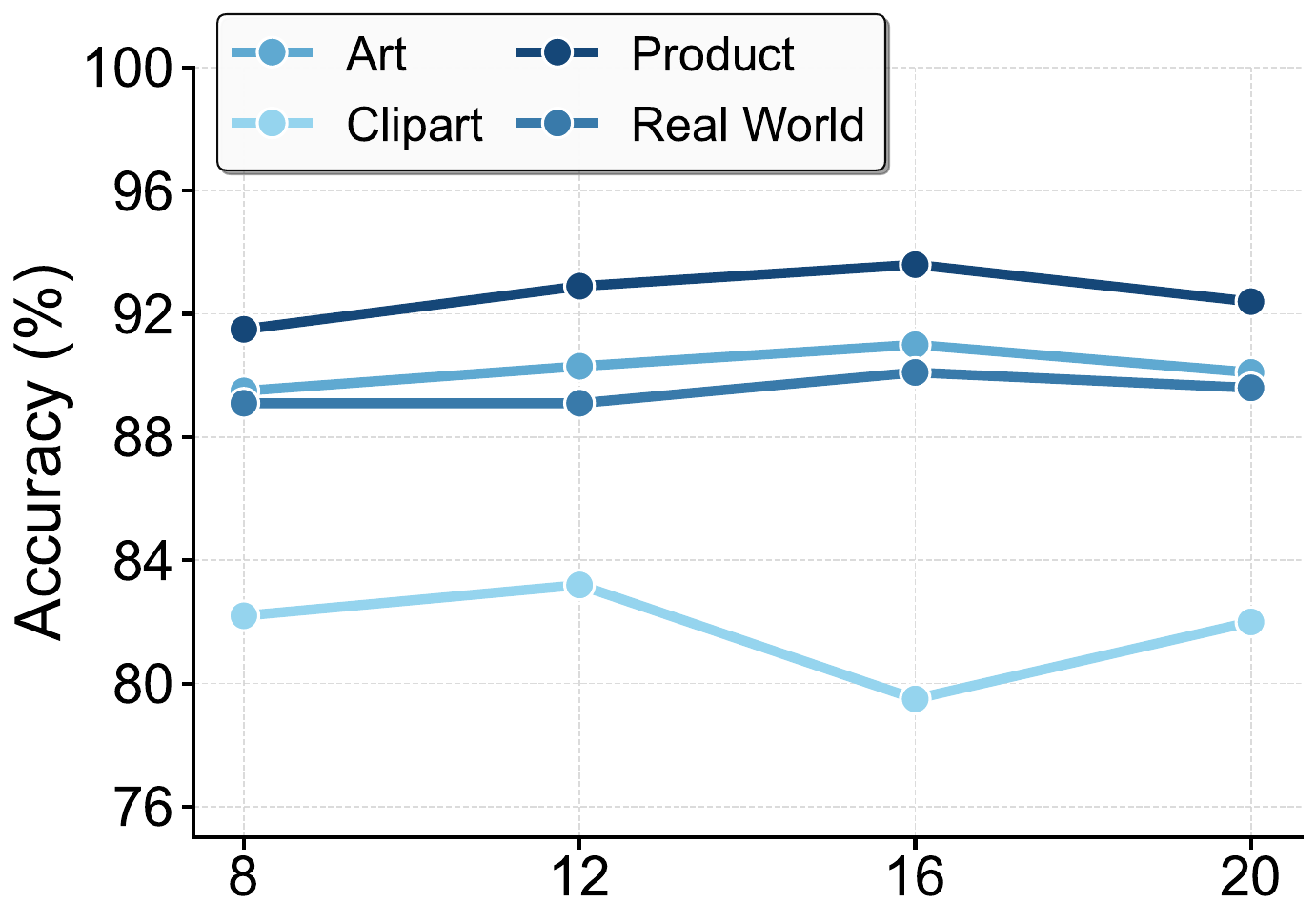}%
    \label{fig:adapter_m1_m2}
}
\hfill
\subfloat[Ablation study on $r_2$]{%
    \includegraphics[width=0.24\linewidth]{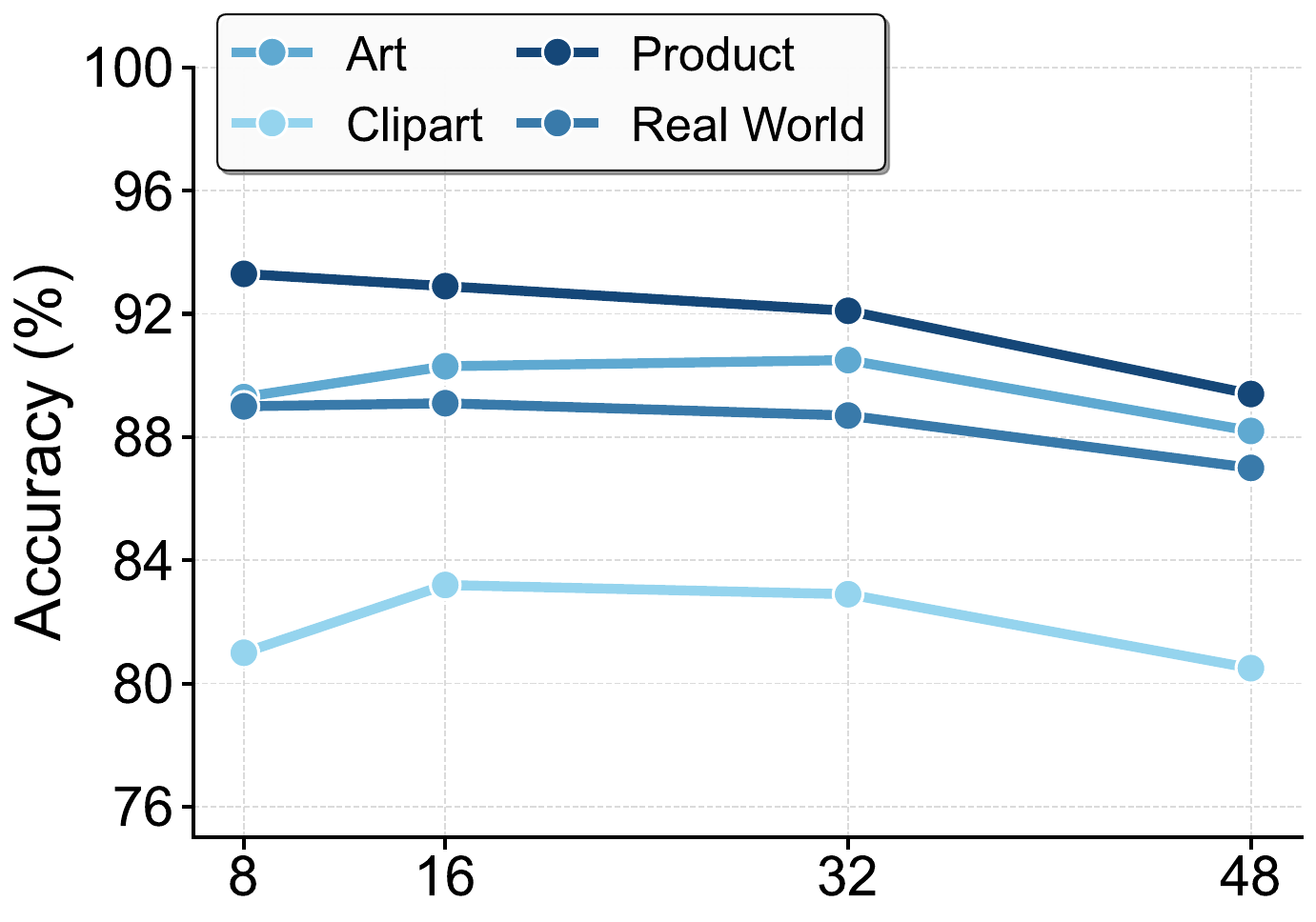}%
    \label{fig:adapter_r}
}
\caption{\method~- Adapter ablation studies. (a) threshold parameter $\beta$ (b) threshold parameter $\tau$ (c) prompt length $M_1$ and $M_2$ (d) inner dimension $r_2$.}
\vspace{-1ex}
\label{fig:ablation_adapter}
\end{figure*}

We also perform an ablation study to analyze the effect of the number of clusters, with the results summarized in Table~\ref{tab:cluster_ablation}. As shown, using three clusters consistently yields the best overall performance across all PEFT modules. However, we note that the exact number of clusters is not a critical hyperparameter, as the performance variations remain within a narrow 1\% margin, indicating the robustness of our method to this choice. Interestingly, a general downward trend in performance is observed as the number of clusters increases. This suggests that using too many learning cycles may lead to over-fragmentation of the data, causing the model to overfit to small, isolated subsets of samples.

% \vspace{1ex}
Overall, these comprehensive experiments highlight the clear advantage of learning from a structured, easy-to-hard curriculum rather than treating all target samples uniformly. The consistent gains across different modules and datasets demonstrate that our “learn, refine, and rehearse” cycle not only stabilizes the training process but also enhances generalization across diverse and challenging domain shifts.

\subsection{Analysis on Source Pretraining}

To validate the impact of our source pretraining strategy, we conduct a targeted ablation study comparing the performance of all three PEFT modules with and without this initial step. The results, detailed in Figure~\ref{fig:combined_source_pretrain}, reveal that this pretraining is crucial for achieving optimal performance.

As shown, its inclusion consistently yields higher accuracy across all modules, with substantial performance gains of 2.4\% for Visual Prompt, 2.3\% for LoRA, and 1.7\% for Adapter. Unlike the progressive alignment strategy, source pretraining provides a universal performance lift across all domains. This demonstrates that leveraging supervised signals from the source domains enables a more robust initial adaptation and, critically, more reliable target label estimation. The resulting improvement in pseudo-label quality serves as a stronger learning signal during the early stages of training, leading to more stable convergence and superior downstream performance. This confirms that our simple source pretraining strategy provides a valuable foundation for generating high-confidence pseudo-labels, which are crucial for the success of our progressive learning strategy.

\subsection{Hyperparameter Ablation Study}
A comprehensive analysis of key hyperparameters is conducted to evaluate the sensitivity and robustness of our framework. The experimental results are summarized in Figure~\ref{fig:ablation_prompt} for Visual Prompt, Figure~\ref{fig:ablation_lora} for LoRA, and Figure~\ref{fig:ablation_adapter} for Adapter, respectively. We investigate the effect of four key factors: the rehearse threshold $\beta$, the initial pseudo-label selection threshold $\tau$, the textual prompt lengths $M_1$ and $M_2$, and PEFT-module-specific parameters.

We first analyze the impact of the initial pseudo-label selection threshold $\tau$ on model performance. As shown, the best results for all PEFT modules are obtained at $\tau = 0.6$. We observe a clear trend where performance improves as $\tau$ increases from 0.4 to 0.6, followed by a slight decline when $\tau$ reaches 0.7. Notably, while Visual Prompt and LoRA remain relatively stable across different $\tau$ values, we find larger fluctuations for Adapter, suggesting that its optimization is more sensitive to the pseudo-label quality.

Next, we examine the rehearse threshold $\beta$. Performance remains relatively stable for $\beta$ values of 0.6, 0.7, and 0.8 across all PEFT modules. However, at $\beta = 0.9$, performance drops sharply. This is because such a high threshold filters out too many samples for subsequent stages, causing catastrophic forgetting of previously learned knowledge. For smaller $\beta$ values, while more samples are retained, the additional training data does not lead to notable performance improvements but requires greater computational resources. We find that $\beta = 0.8$ provides the best trade-off between performance and efficiency. This is particularly important for large-scale datasets such as DomainNet, where the initial $\tau$ is set to 0.4. Our experiments show that this setting improves training efficiency by approximately 30\%.

We then study the influence of the textual prompt lengths $M_1$ and $M_2$, where we set $M_1 = M_2$ for simplicity. The best performance is achieved at $M_1 = M_2 = 12$. We hypothesize that shorter prompts lack sufficient semantic capacity to guide adaptation effectively, while overly long prompts introduce redundancy or noise, which may hinder training. This result highlights the importance of selecting an optimal prompt length to balance expressiveness and regularization.

Finally, we examine module-specific hyperparameters: visual prompt length $M_3$, LoRA rank $r_1$, and Adapter inner dimension $r_2$. Both $M_3$ and $r_1$ yield stable performance across tested values, with peak performance at $M_3 = 20$ and $r_1 = 8$. For $r_2$, however, we observe that increasing its value tends to decrease overall accuracy, and thus we choose $r_2=16$. This finding once again underscores that the Adapter module, while powerful, presents greater optimization challenges and is more sensitive to its hyperparameter configuration.

\section{Conclusion}

In this paper, we presented \textbf{\method}, a progressive framework for multi-source unsupervised domain adaptation that adapts CLIP through a structured “\textit{learn, refine, and rehearse}” cycle. By organizing the adaptation process using balanced class clustering and gradually expanding from easy to hard samples, \method~improves training stability and robustness. Specifically, during the learning phase, we train domain-specific textual prompts and a shared visual PEFT module, aligning prompts through a denoising autoencoder with alignment regularization. In the refine-and-rehearse phase, confident samples from previous stages are refined with the updated model and reused to guide training in subsequent stages, enabling efficient knowledge transfer. Furthermore, to enhance pseudo-label quality, we ensemble source-pretrained CLIP models for more reliable target supervision. Experiments on ImageCLEF, Office-Home, and DomainNet show that \method~achieves state-of-the-art performance across all benchmarks. These results highlight the effectiveness of progressive alignment and parameter-efficient adaptation in vision-language models under domain shift.

% \section*{Acknowledgments}
% This should be a simple paragraph before the References to thank those individuals and institutions who have supported your work on this article.

%\begin{thebibliography}{1}
\bibliographystyle{IEEEtran}
\bibliography{refs}

% \newpage

\section{Biography Section}

\begin{IEEEbiography}[{\includegraphics[width=1.2in,height=1.25in,clip,keepaspectratio]{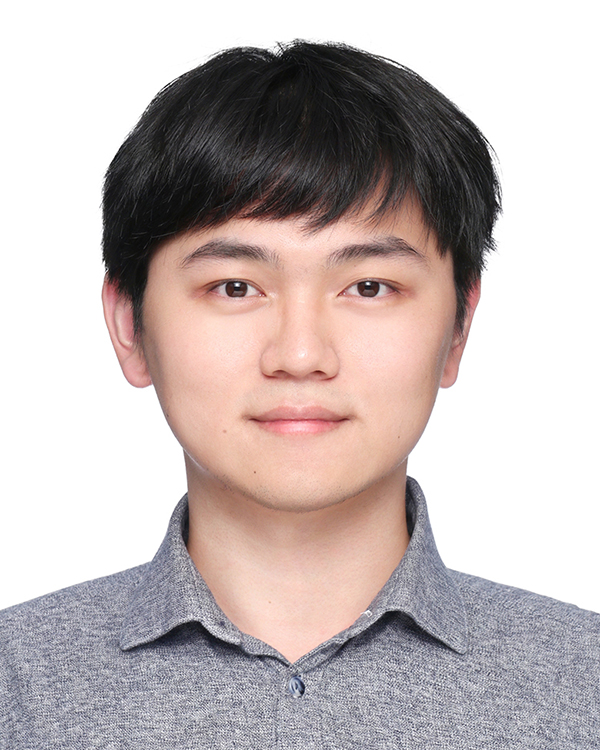}}]{Haoran Chen} 
is currently working toward the Ph.D.
degree in Computer Science at Fudan University,
advised by Prof. Zuxuan Wu and Prof. Yu-Gang Jiang. His research interests include transfer learning, continual learning, and multimodal large language models. \end{IEEEbiography}

\begin{IEEEbiography}[{\includegraphics[width=1.2in,height=1.25in,clip,keepaspectratio]{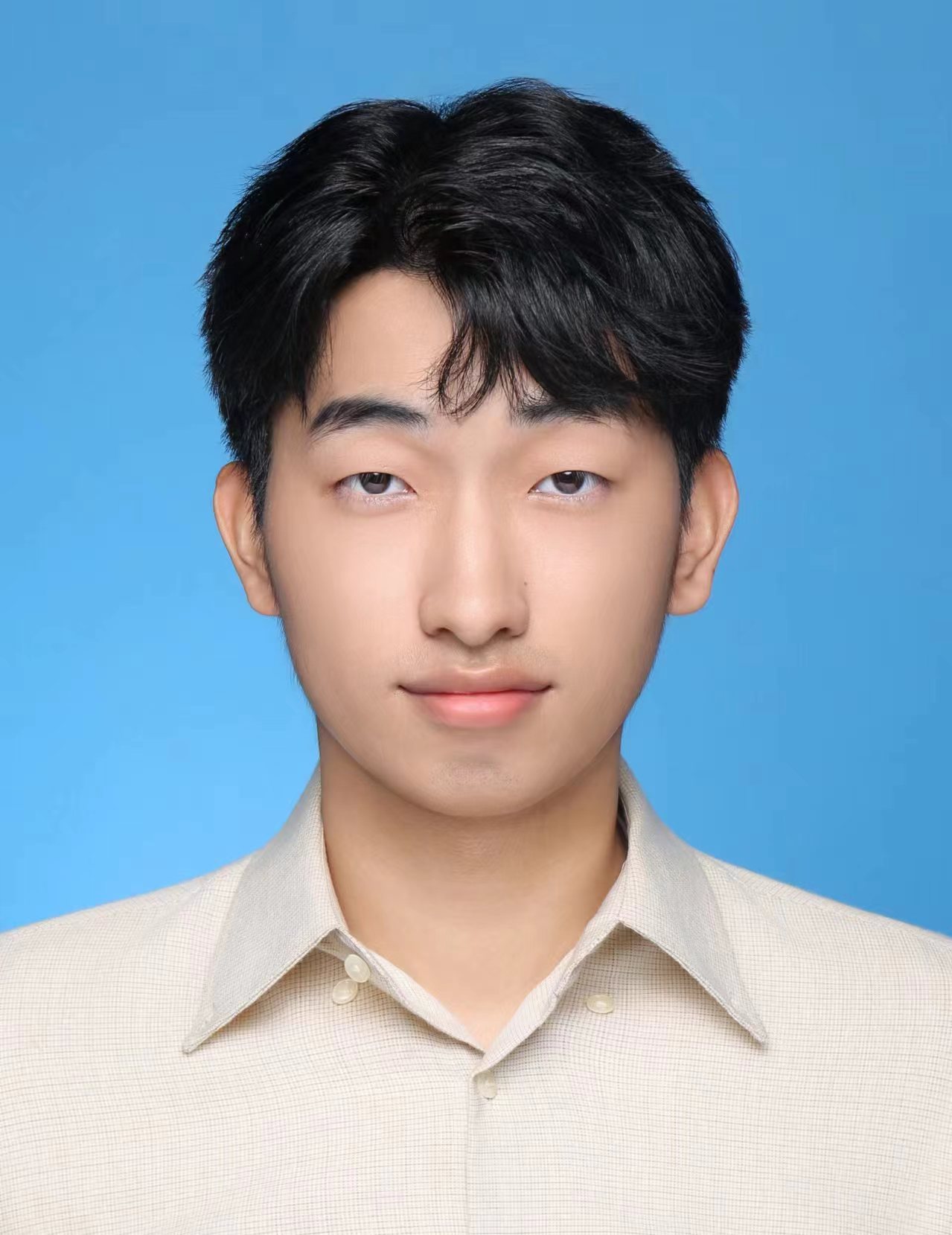}}]{Zexiao Wang} 
is a first year Masters student in Computer Science at Fudan University,
advised by Prof. Zuxuan Wu. His research interests include transfer learning and continual learning. \end{IEEEbiography}

\begin{IEEEbiography}[{\includegraphics[width=1.2in,height=1.25in,clip,keepaspectratio]{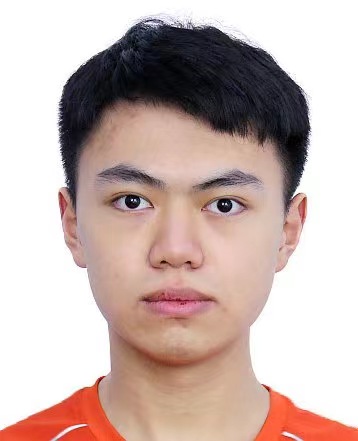}}]{Haidong Cao} 
is a second year Master student in Computer Science at Fudan University,
advised by Prof. Zuxuan Wu. His research interests include computer vision and robot learning. \end{IEEEbiography}

\begin{IEEEbiography}[{\includegraphics[width=1.2in,height=1.25in,clip,keepaspectratio]{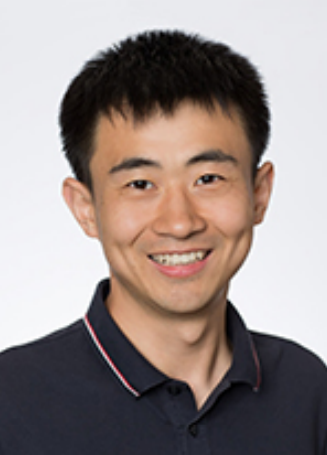}}]{Zuxuan Wu} 
received his Ph.D. in Computer Science from the University of Maryland with Prof. Larry Davis in 2020. He is currently an Associate Professor at the Institute of Trustworthy Embodied AI, Fudan University. His research interests are in computer vision and deep learning. His work has been recognized by an AI 2000 Most Influential Scholars Honorable Mention in 2021, a Microsoft Research PhD Fellowship (10 people Worldwide) in 2019 and a Snap PhD Fellowship (10 people Worldwide) in 2017. \end{IEEEbiography}

\begin{IEEEbiography}[{\includegraphics[width=1.2in,height=1.25in,clip,keepaspectratio]{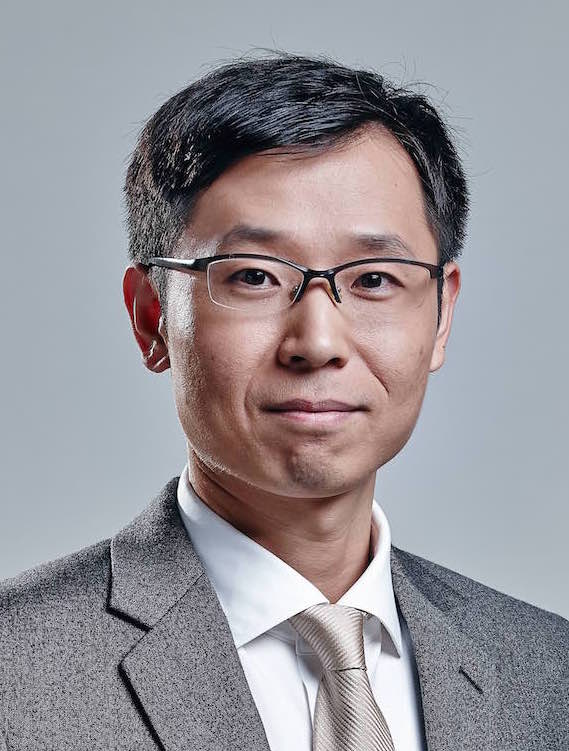}}]{Yu-Gang Jiang} (Fellow, IEEE) received the PhD degree in Computer Science from City University of Hong Kong in 2009 and worked as a Postdoctoral Research Scientist at Columbia University, New York, during 2009-2011. He is currently a Distinguished Professor at the Institute of Trustworthy Embodied AI, Fudan University, Shanghai, China. His research lies in the areas of multimedia, computer vision, embodied AI and trustworthy AI. His research has led to the development of innovative AI tools that have been used in many practical applications like defect detection for high-speed railway infrastructures. His open-source video analysis toolkits and datasets such as CU-VIREO374, CCV, THUMOS, FCVID and WildDeepfake have been widely used in both academia and industry. He currently serves as Chair of ACM Shanghai Chapter and Associate Editor of several international journals. For contributions to large-scale and trustworthy video analysis, he was elected to Fellow of IEEE, IAPR and CCF.
 \end{IEEEbiography}

\end{document}